 \documentclass[nopreprintline, final,5p,times,twocolumn,authoryear]{elsarticle}

\usepackage{amssymb}
\usepackage{amsthm}
\usepackage{amsmath,amssymb,amsfonts}
\usepackage{lipsum}
\usepackage{setspace}
\usepackage{algorithm}
\usepackage{algpseudocode}
\usepackage{arydshln}
\usepackage{url}
\algnewcommand{\LeftComment}[1]{\Statex \(\triangleright\) #1}

\journal{}

\begin{document}

\begin{frontmatter}



\title{IterMask3D: Unsupervised Anomaly Detection and Segmentation with Test-Time Iterative Mask Refinement in 3D Brain MRI}


\author[1]{Ziyun {Liang}}
\author[1,2]{Xiaoqing {Guo}}
\author[1]{Wentian {Xu}}
\author[1]{Yasin {Ibrahim}}
\author[3]{Natalie {Voets}}
\author[4]{Pieter M {Pretorius}}
\author[1]{J. Alison {Noble}}
\author[1]{Konstantinos {Kamnitsas}\corref{cor1}}
\cortext[cor1]{Corresponding author.}
\affiliation[1]{Department of Engineering Science, University of Oxford, UK}
\affiliation[2]{Department of Computer Science, Hong Kong Baptist University}
\affiliation[3]{Nuffield Department of Clinical Neurosciences, University of Oxford}
\affiliation[4]{Department of Neuroradiology, Oxford University Hospitals NHS Foundation Trust}

\begin{abstract}
Unsupervised anomaly detection and segmentation methods train a model to learn the training distribution as `normal'. In the testing phase, they identify patterns that deviate from this normal distribution as `anomalies'. 
To learn the `normal' distribution, prevailing methods 
 corrupt the images and train a model to reconstruct them. 
During testing, the model attempts to reconstruct corrupted inputs based on the learned `normal' distribution. 
Deviations from this distribution lead to high reconstruction errors, which indicate potential anomalies. 
However, corrupting an input image inevitably causes information loss even in normal regions, leading to suboptimal reconstruction and an increased risk of false positives.
To alleviate this, we propose $\rm{IterMask3D}$, an iterative spatial mask-refining strategy designed for 3D brain MRI. 
We iteratively spatially mask areas of the image as corruption and reconstruct them, then shrink the mask based on reconstruction error. This process iteratively unmasks `normal' areas to the model, whose information further guides reconstruction of `normal' patterns under the mask to be reconstructed accurately, reducing false positives. 
In addition, to achieve better reconstruction performance, we also propose using high-frequency image content as additional structural information to guide the reconstruction of the masked area.
Extensive experiments on the detection of both synthetic and real-world imaging artifacts, as well as segmentation of various pathological lesions across multiple MRI sequences, consistently demonstrate the effectiveness of our proposed method. Code is available at https://github.com/ZiyunLiang/IterMask3D. 
\end{abstract}



\begin{keyword}
Unsupervised Anomaly Segmentation \sep 3D Brain MRI \sep Anomaly Detection  



\end{keyword}

\end{frontmatter}




\section{Introduction}
\label{sec:introduction}

Segmenting anomalies is crucial in the field of medical image analysis as it enables applications such as early disease detection and diagnosis, guides treatment planning, and reduces clinical workload.
Conventional anomaly segmentation methods are mostly supervised, relying on annotated training data, where images contain anomalies with corresponding manual labels. 
Trained on a limited set of data types, the model can only segment anomalies resembling those in its training data, and struggles to detect other types of unseen anomalies. In the context of brain MRI images, the focus of this study, this type of methods typically target a specific pathology \citep{kamnitsas2017efficient, isensee2021nnu, stollenga2015parallel}. 
Unsupervised anomaly segmentation, on the other hand, does not require any `anomalous' images or their manual segmentations during training. Instead, it is trained exclusively on `normal' images and treats this training distribution as the `normal' reference. During testing, the method regards any deviation from this reference as `anomaly' and attempts to segment it. This enables the detection of previously unseen patterns, including new pathologies or even imaging artifacts, making it suitable for tasks such as automatic screening for incidental findings \citep{rowley2023incidental} or automatic scan quality control \citep{hendriks2024systematic}. Such an approach is especially beneficial when training data are limited and the specific pathology type is unknown in advance, facilitating the detection of rare diseases.
In this work, we explore unsupervised anomaly segmentation, with a focus on brain MRI data. 

\begin{figure}[!t]
  \centering
\includegraphics[width=\columnwidth,height=0.9\textheight,keepaspectratio]{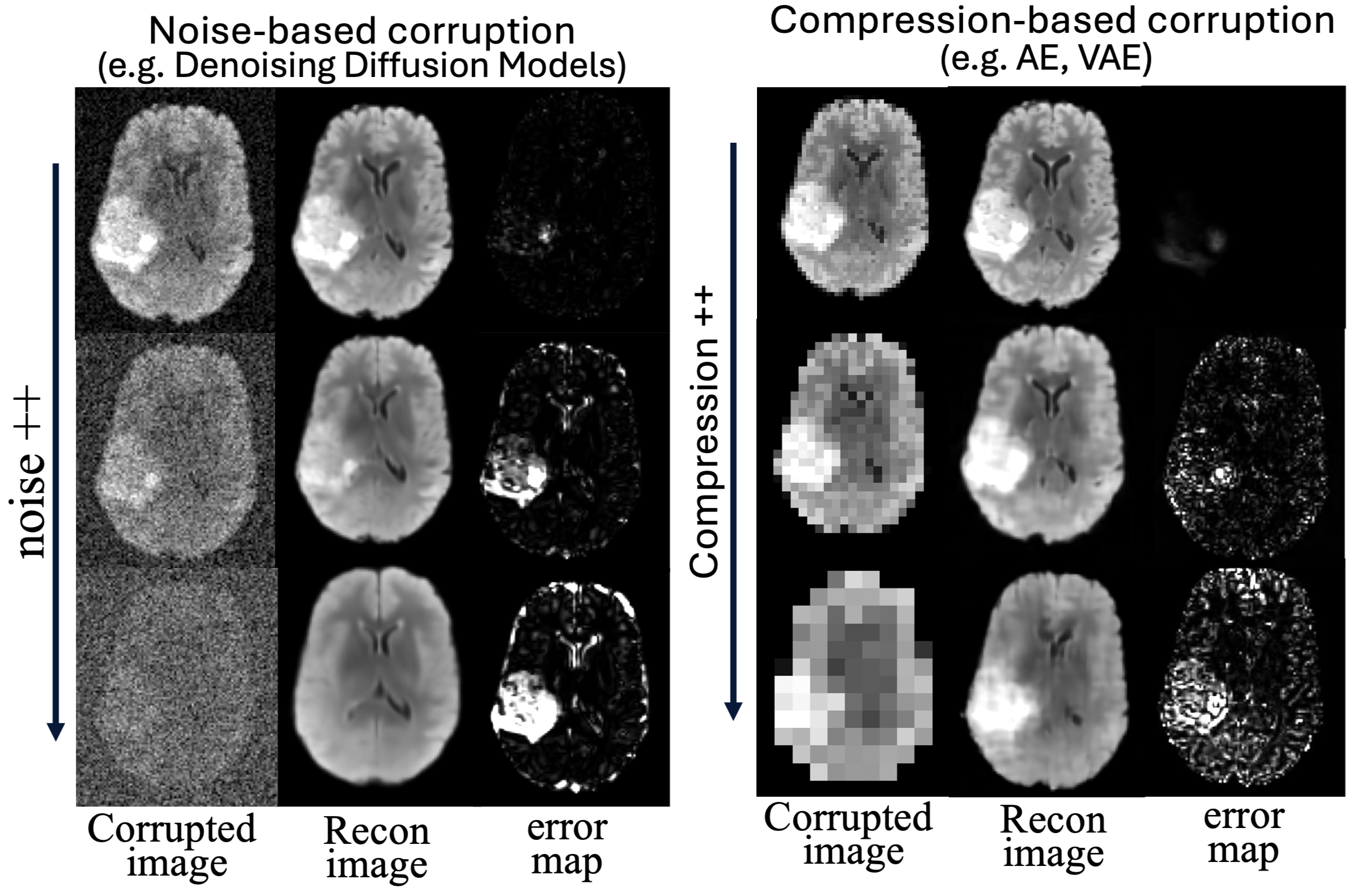}
  \caption{\textbf{Sensitivity–Precision Trade-off} in reconstruction-based methods: When corruption (e.g., noise or compression) is lower, the reconstruction error over anomalous regions (the hyper-intense tumor) remains low, often resulting in missed detections. As corruption increases, errors in abnormal regions become more pronounced, improving sensitivity—but at the cost of higher errors in normal regions, which can increase false positives and reduce precision.}
  \label{fig:intro}
\end{figure}

A prevalent strategy for learning the training distribution is reconstruction-based methods that follow the `corrupt and reconstruct' paradigm. During training, images are first artificially corrupted, then models are trained to reconstruct the provided `normal' in-distribution images, thus internalizing the normal distribution through the reconstruction process. At testing, regions that deviate from the learned normal distribution, i.e., anomalies, tend to yield high reconstruction errors, which can then be exploited for segmentation. 
Common ways of adding corruptions include compression \citep{atlason2019unsupervised,baid_rsna-asnr-miccai_2021} or noise as in diffusion models \citep{pinaya2022fast,bercea2023mask,wolleb2022diffusion}, where the model is tasked to reconstruct from these corruptions. A common challenge of these types of methods is \textit{sensitivity-precision trade-off}, as shown in Fig.~\ref{fig:intro}. In order to amplify reconstruction error in abnormal areas for better segmentation, more corruption is preferred. However, since the location of anomalies is unknown in advance, corruption is applied to the entire image. Thus, more corruptions will not only increase the reconstruction error in abnormal area, but it will also increase it in normal areas, which leads to false positives. As shown in Fig.~\ref{fig:intro}, increasing the noise level and compression results in higher reconstruction error over the anomaly (in this case, the hyper-intense tumor), thus leads to better detection of the anomaly area. But at the same time, normal areas also have larger reconstruction errors, leading to false positives.
This \textit{sensitivity-precision trade-off} serves as the main motivation of our paper: \textit{Can we use input corruptions to amplify reconstruction errors in anomalous areas for better segmentation, while preserving the information in normal areas to minimize reconstruction errors and reduce false positives?}


To address this problem, we propose $IterMask3D$, an iterative mask refinement method for unsupervised anomaly segmentation and detection that can be used on 3D MRI data. To amplify reconstruction errors in anomalous regions, we introduce spatial masking as a form of corruption. To minimize errors in normal regions, we employ an iterative spatial mask shrinking process, gradually refining the mask toward the anomaly to reduce information loss. As normal tissues are progressively unmasked and fed into the model, they guide the reconstruction of normal structures beneath the mask to reduce false positives. Additionally, we generate a structural image using high-frequency masking, providing structural guidance to steer the reconstruction of normal areas for the mask-shrinking process. 

Our main contributions are:

(1) We propose a 3D unsupervised anomaly segmentation and detection method $IterMask3D$, which iteratively shrinks a spatial mask towards the anomaly to reduce false positives on normal areas. 

(2) We further propose the use of frequency masking to generate high-frequency images as structural guidance for the mask-shrinking process, enhancing the reconstruction of normal areas and further reducing false positives.

(3) We introduce a subject-specific thresholding strategy that determines, at each iteration, which regions on the error map can be confidently classified as normal. This allows the iterative mask-shrinking process to automatically determine the optimal stopping point for each individual subject without the need to require extra information from validation data.

(4) We evaluate the effectiveness of our method on diverse types of anomalies, including both pathological and artifact-based abnormalities. Specifically, we test our model on synthetic and real-world artifacts, as well on tumor and stroke as pathologies across multiple MRI sequences.

This paper builds upon our previous work \citep{liang2024itermask}, where we introduced Iterative Mask Refinement ($IterMask^2$) for unsupervised anomaly segmentation in 2D brain MRI pathology. The 2D-based model follows the standard practice of anomaly segmentation, where it is trained on healthy slices and evaluated on anomaly slices from the same dataset. This builds upon it by: (a) extending the method to 3D and evaluate it on real-world training and testing data from different distributions, further advancing its applicability to real-world settings. (b) We extend our evaluation beyond pathological anomalies to include imaging artifacts, assessing the model's effectiveness in detecting both types of anomalies.
(c) Instead of relying on a fixed threshold for the entire dataset, we adopt an image-specific thresholding strategy that automatically determines when to stop the mask-shrinking process—without requiring access to any validation data. 
(d) In the last work, our method required two separate models: one for the initial prediction when masking the whole brain area and another for the subsequent mask-shrinking process. In this work, we unify these steps into a single model for the entire mask-shrinking process. (e) We extend the mask refinement process to support both shrinking and expansion, allowing more flexible and being able to correct the mistakes from previous steps. 

The rest of the paper is organized as follows. The related works are described in Section~\ref{sec:relatedwork}. Our method $IterMask3D$ is introduced in Section~\ref{sec:methods}.
The experimental settings are introduced in Section~\ref{sec:experiments_setting}. The experimental results are presented in Section~\ref{sec:experiments}, covering four tasks: detection of synthetic artifacts (Subsection~\ref{subsec:synthetic_artifact}), detection of real-world artifacts (Subsection~\ref{subsec:real_artifact}), pathology segmentation on 2D data (Subsection~\ref{subsec:2d_patho}), and pathology segmentation on 3D data (Subsection~\ref{subsec:3d_patho}).
Finally, the discussion and conclusion are presented in Section~\ref{sec:discu&conclu}.

\section{Related Work}
\label{sec:relatedwork}

\textbf{Reconstruction-based unsupervised anomaly segmentation:} These approaches are among the most prevalent for unsupervised anomaly segmentation, typically following a `corrupt and reconstruct' paradigm (as outlined in Section~\ref{sec:introduction}). The core idea is to learn the distribution of normal data by training a model to reconstruct inputs after they have been deliberately corrupted. One way to achieve this corruption is via compression, as used by Autoencoders (AE) \citep{atlason2019unsupervised,baur2021autoencoders,cai2024rethinking}, Variational Autoencoders (VAE) \citep{baur2021autoencoders,pawlowski_unsupervised_2018,shen_unsupervised_2019}, and Vector Quantized Variational Autoencoders (VQ-VAE) \citep{pinaya2022unsupervised}. Another common strategy, generative models, employs noise as corruption. 
Among these methods, f-AnoGAN \citep{schlegl2019f} builds on Generative Adversarial Networks (GANs), but the notoriously unstable training of GANs often limits their practical applicability. Recently, diffusion models \citep{ho2020denoising} have shown promise across diverse medical imaging tasks \citep{wu2024medsegdiff,khader2023denoising,zheng2024deformation}, with applications extending to unsupervised anomaly segmentation \citep{bercea2023mask,wolleb2022diffusion,pinaya2022fast}. In these approaches, inputs are typically corrupted with a certain level of Gaussian noise \citep{pinaya2022fast,bercea2023mask} before being reconstructed.

A key challenge in reconstruction-based methods is determining how much corruption (e.g. compression or noise) to inject. Greater corruption can amplify reconstruction errors in anomalous regions but also in normal areas, increasing false positives \citep{bercea2023aes}. We refer to this balance between capturing anomalies (sensitivity) and preserving normal regions (precision) as the \emph{sensitivity-precision trade-off} (Fig.~\ref{fig:intro}). One of our primary objectives is to alleviate this trade-off by ensuring that reconstruction errors remain high for anomalies and low for normal areas. 
Example of effort made in this direction is the work of \citep{bercea2023mask} that proposes an iterative process where a mask covering all anomalies is generated first, and then the masked region is reconstructed using a diffusion model. Although similar in its iterative design, this approach does not refine the potentially suboptimal initial mask, nor does it leverage auxiliary guidance to accurately reconstruct the masked content. As a result, it can struggle to faithfully recover normal tissues. Our method overcomes these issues more effectively, as we will demonstrate.

\textbf{Non-reconstruction-based unsupervised anomaly segmentation:}
A transformer-based approach is introduced in \citep{pinaya2022unsupervised}, where a VQ-VAE serves as the foundational architecture. The method enhances the standard reconstruction-based framework by incorporating a transformer as an auto-regressive model operating on the compressed latent representations.
This work was applied to 3D brain MRI for anomaly detection, using only `healthy' subjects for training as normal and subsequently identifying pathologies as anomalies. However, the transformer component requires an ensemble prediction of differently-ordered latent representations, leading to inefficient computation. 

Recently, cross-sequence translation has been proposed for reconstruction \citep{liang2023modality}. In this approach, a model is trained to translate normal tissues between different MRI sequences, with the assumption that anomalies will fail to be translated correctly and can thus be segmented. However, this method requires multi-modal data and may suffer from information loss if features are not fully shared across sequences, potentially leading to false positives.

Another type of method that shows promising performance is the denoising-based methods \citep{kascenas_denoising_2022}. Unlike `corrupt and reconstruct', this adds noise to inputs only during \textit{training} (not testing), then learns to remove it. During testing, anomalies are treated as noise and removed. A key limitation, however, is that the noise added during training must approximate the appearance of anomalies, requiring prior knowledge of their morphology. This constraint is impractical when aiming to detect all pathologies, as we do in this work.

\section{Methods}
\label{sec:methods}

\begin{figure*}[t!] 
    \centering
    \includegraphics[width=\textwidth]{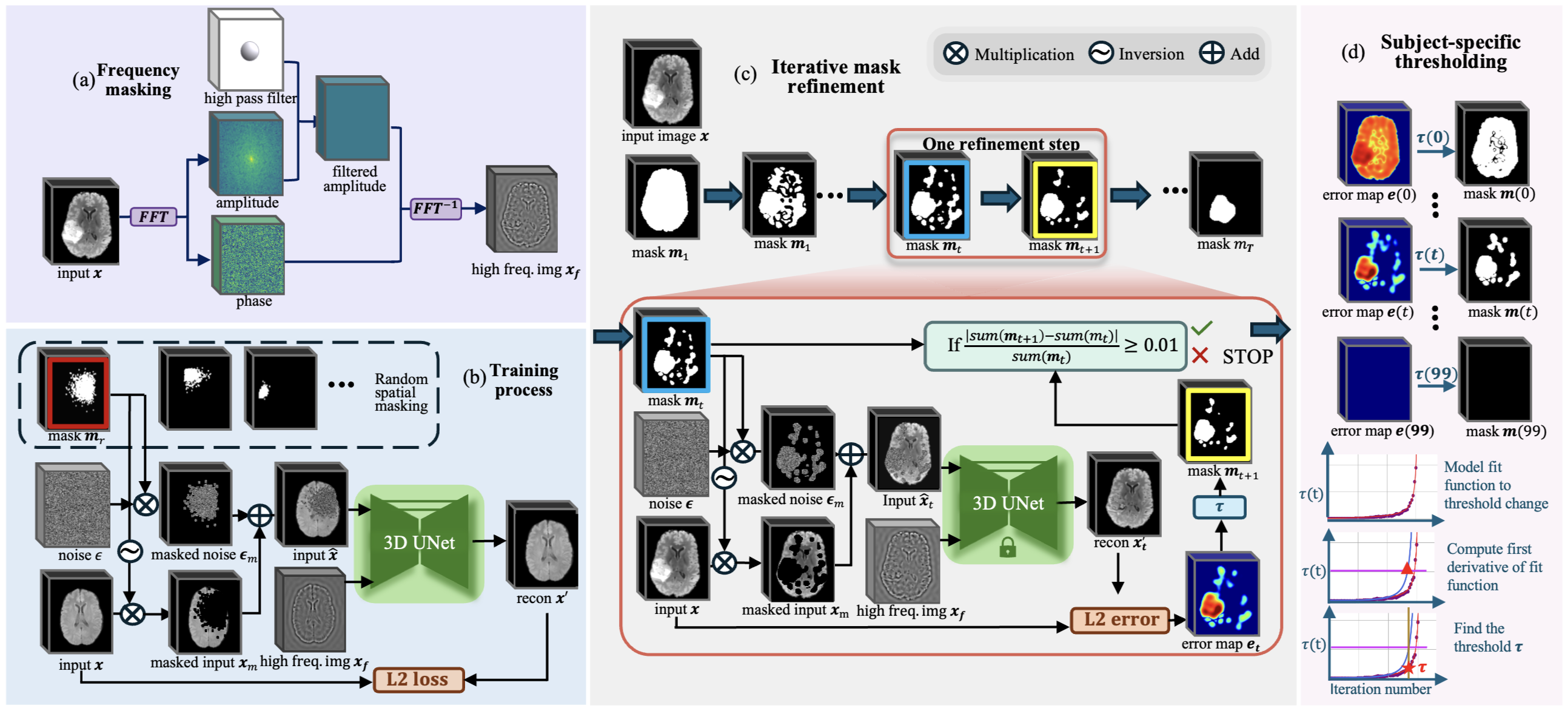} 
    \caption{
     \textbf{Overview of the proposed approach.}
(a) Frequency-based masking: Extract structural information $x_f$ with a high-pass filter to isolate the high-frequency components of the image. 
(b) Training process: Input image $x$ is masked using random spatial masking, and the model learns to reconstruct the masked area with $x_f$ generated in (a) as an auxiliary input. 
(c) Iterative mask refinement: Iterative refinement of the spatial mask, where the mask $m_t$ at iteration $t$ gradually shrinks towards the anomalous region. This refinement is guided by the spatially unmasked portions of the image $x_m$ and $x_f$ from (a). (d) Subject-Specific Thresholding: This component illustrates how the threshold $\tau$ is determined for iterative mask refinement on a per-sample basis. The mask is progressively shrunk by 1\% of the brain volume at each time step $t$ until it fully disappears. We model $\tau(t)$ using a fitted function, and compute its derivative to identify the point where the threshold begins to change abruptly. This point is selected as the sample-specific threshold $\tau$ for the current sample. }
    \label{fig:main}
\end{figure*}

\subsection{The overview}
\label{sec:metoverview}

Our proposed method is a reconstruction-based unsupervised anomaly segmentation method. The training dataset, $D_{tr}=\left\{\mathbf{x}^{i}\right\}^N_{i=1}$, consists of `normal' in-distribution samples. The model aims to learn the training distribution following the `corrupt and reconstruct' paradigm as previously introduced in Section~\ref{sec:introduction}, which learns to reconstruct in-distribution samples after corruption. 
During testing, given testing dataset $D_{te}=\left\{\mathbf{x}^{i}\right\}^M_{i=1}$, the model aims to segment any patterns outside the training distribution (unseen during training) as anomalies. 
To mitigate the sensitivity-precision trade-off in reconstruction-based methods, we propose an iterative spatial mask refinement process, $IterMask3D$ as shown in Fig.~\ref{fig:main} (c) \textit{Testing Process}.
To ensure large reconstruction errors in abnormal regions for effective anomaly segmentation, spatial masking is applied to completely mask anomalies before reconstruction.
Additionally, to accurately reconstruct normal areas with minimal reconstruction error, we gradually shrink the mask towards the anomaly. This process progressively unmasks normal regions, providing the network with more information on case-specific `normal' patterns to better reconstruct normal areas under the mask. 

Specifically, as shown in Fig.~\ref{fig:main} (c) \textit{Iterative Mask Refinement},
we start by masking the entire brain area. In each iteration, the network reconstructs the masked regions and areas with small errors are identified as normal and subsequently unmasked. With every iteration, more information is revealed, allowing the network to refine its reconstruction. 

However, as the network is guided only by the unmasked regions, it must infer the missing structures in a generative manner. Incorrectly inferred structures will introduce reconstruction errors even in normal areas, leading to false positives. 
To address this, we further propose incorporating high-frequency information (achieved via low-frequency masking in the Fourier space) to guide the structural reconstruction of masked regions, as shown in Fig.~\ref{fig:main} (a) 
In each iteration, the threshold for identifying normal regions is decided by the \textit{Subject-specific thresholding} strategy, as shown in Fig.~\ref{fig:main} (d). 
As the mask is progressively shrunk by a fixed proportion, the threshold is selected based on the characteristic change in threshold values throughout the mask-shrinking process.

Finally, the training process is depicted in Fig.~\ref{fig:main} (b). 
To train the reconstruction model, we apply randomly generated Gaussian masks as spatial masks to the input image and combine them with input of high-frequency components (generated as shown in Fig.\ref{fig:main}(a)). The model is trained to reconstruct the masked regions using this combined input. 
In the following sections, we introduce Iterative Spatial Mask Refinement (Fig.\ref{fig:main} (c)) in Section \ref{sec:metiter}, frequency masking (Fig.\ref{fig:main} (a)) in Section \ref{sec:metfreq}, subject-specific thresholding strategy (Fig.\ref{fig:main} (d)) in Section \ref{sec:thres}, and the training process (Fig.\ref{fig:main} (b)) in Section \ref{sec:train}.

\subsection{Iterative Spatial Mask Refinement for Testing}
\label{sec:metiter}

In this section, we provide a detailed explanation of how our iterative spatial mask refinement method works during testing. The process iteratively refines the spatial mask towards the anomaly as shown in the top part of Fig.~\ref{fig:main} (c). We define the iteration number as $t$, where $t=0,1,...,T$. In the first iteration ($t=0$), given an input test image $\mathbf{x}$, the spatial mask starts shrinking from an initial mask covering the entire brain $\textbf{m}_0$. 
The mask is then progressively shrunk to mask $\textbf{m}_t$ at iteration $t$, narrowing its focus towards the anomaly, until the final mask reaches $\textbf{m}_T$. 

To describe the process in greater detail, we focus on one iteration $t$ extracted from the complete mask shrinking process as shown in the bottom part of Fig.~\ref{fig:main} (c). The input to this iteration is original test image $\mathbf{x}$ and mask $\textbf{m}_t$. Gaussian noise $\boldsymbol{\epsilon}$ is generated with the same dimensions as the original test image, and it is used to cover the masked area in the image $\boldsymbol{\epsilon_{m}}$ (Eq.~\ref{equ1}). The unmasked area is already confidently identified as normal from previous iterations so copied directly from the input image $\mathbf{x}_m$ (Eq.~\ref{equ2}). Then they are added together as the input to the model at iteration $t$ as $\hat{\mathbf{x}}_t$ (Eq.~\ref{equ3}):

\begin{align}
\boldsymbol{\epsilon_{m}}=\textbf{m}_t \odot \boldsymbol{\epsilon} \label{equ1}\\
\mathbf{x}_m = \mathbf{x} \odot (1-\mathbf{m}_t) \label{equ2}\\
\hat{\mathbf{x}}_t = \boldsymbol{\epsilon_{m}} +  \mathbf{x}_m \label{equ3}
\end{align}
Here, $\odot$ represents element-wise multiplication. Then, the masked image $\hat{\mathbf{x}}_t$ and high frequency image $\mathbf{x}_f$ is input to our model $f$ to reconstruct the spatially masked areas. The model outputs reconstructed prediction $\mathbf{x}_t'$: 
\begin{align}
\mathbf{x}_t' = f(\hat{\mathbf{x}}_t, \mathbf{x}_f) \label{equ4}
\end{align}

The high-frequency image $\mathbf{x}_f$ here is an additional condition to the model, which will be discussed in Section \ref{sec:metfreq}.
After the model reconstructs the masked area, we compute the $L_2$ distance as an error map between the reconstructed image $\mathbf{x}_t'$ and the original test image $\mathbf{x}$:
\begin{align}
    \mathbf{e}_t = \lVert \mathbf{x}_t' - \mathbf{x} \rVert _2\
\end{align}
Then, on the error map, we decide where is confidently identified as normal in the current iteration by comparing the error map $\mathbf{e}_t$ with a threshold $\tau_{stop}$.
Pixels with error that are below the specified threshold, $\tau_{stop}$, will be removed from the masked area, and the mask will shrink accordingly (Eq.~\ref{eqa:mt}). A detailed analysis on selecting the optimal $\tau_{stop}$ for current sample can be found in Section \ref{sec:thres}.
\begin{align}
    \mathbf{m}_{t+1}(i, j) =
    \begin{cases} 
        0, & \text{if } \mathbf{e}_t(i, j) < \tau_{stop} \\
        1, & \text{otherwise}
    \end{cases}
    \label{eqa:mt}
\end{align}
When most of the spatially masked areas are anomalies, introducing more information will not reduce the high error in the anomalous areas and the mask will stop shrinking (shrink less than 1\%), which is how we terminate the testing process.

\begin{algorithm}
\setstretch{1.1}
    \caption{Iterative Spatial Mask Refinement (IterMask3D)}
    \begin{algorithmic}[1]
    \Require 3D Brain MRI (\(\mathbf{x} \in D_{te}\)), High-Frequency Image (\(\textbf{x}_f\)), Subject-specific Threshold (\(\tau_{stop}\)) (Section~\ref{sec:thres})
    \Ensure Anomaly Mask (\(\textbf{m}_T\))
    \State \(m_0 = \mathbf{1}, \quad t = 0\)
    
    \While{mask is still shrinking i.e. {\(\frac{| \sum (\textbf{m}_{t+1}) - \sum (\textbf{m}_t) |}{\sum (\textbf{m}_t)} < 0.01\)}}
        \State \(\boldsymbol{\epsilon} \sim \mathcal{N}(0,1)\)
        \LeftComment Add noise to the masked area
        \State \(\boldsymbol{\epsilon}_m = \textbf{m}_t \odot \boldsymbol{\epsilon}\)
        \LeftComment Leave the unmasked area unchanged
        \State \(\textbf{x}_m = \textbf{x} \odot (1 - \textbf{m}_t)\) 
        \LeftComment Combine these two to obtain a new image
        \State \(\hat{\textbf{x}} = \boldsymbol{\epsilon}_m + \textbf{x}_m\)
        \LeftComment Predict based on combined and high-frequency image
        \State \(\textbf{x}_t' = f(\hat{\textbf{x}}, \textbf{x}_f)\)
        \LeftComment Obtain $L_2$ distance error map for masked area
        \State \(\textbf{e}_t = \|\textbf{x}_t' - \textbf{x}\|_2\)
        \LeftComment For each pixel, unmask areas with error smaller than $\tau_{stop}$
        \State \(\textbf{m}_{t+1}(i,j) =
            \begin{cases} 
              0, & \text{if } \textbf{e}_t(i,j) < \tau_{stop} \\
              1, & \text{otherwise}
            \end{cases}\)
        \State \(t = t + 1\)
    \EndWhile
    \State \Return{\(\textbf{m}_T\)}
    \end{algorithmic}
    \label{algo1}
\end{algorithm}

\subsection{Frequency Masking}
\label{sec:metfreq}
In frequency domain analysis, an image is decomposed into its constituent spatial frequencies using transformations such as the Fourier Transform. Low-frequency components correspond to smooth intensity variations and capture intensities of large-scale image features, including the mean image intensity (DC signal). High-frequency components, on the other hand, encode rapid intensity changes, representing fine-grained details such as edges, textures, tissue boundaries, and the fine-grained contours of anatomical structures \citep{chapelle2019}. As part of frequency decomposition, an image can also be characterized by its amplitude and phase components. 
While the amplitude spectrum determines how strongly each frequency contributes to the overall image, the phase spectrum preserves spatial relationships, ensuring that structures and textures are correctly positioned \citep{oppenheim1981importance, oppenheim1979phase}.  

In our model $f$ from the last section, spatial mask reconstruction relies solely on the unmasked area. However, due to the variability in human tissue structure, the model may generate a plausible yet structurally different reconstruction from the actual brain, leading to reconstruction errors even in normal regions. To address this, we introduce a high-frequency image as a conditioning input, providing structural guidance to the model. Since the model has not been trained to reconstruct anomalies from their high-frequency components, this approach reduces reconstruction errors in normal areas while preserving high reconstruction errors in anomalous regions.

To be more specific, when generating high-frequency images, we first apply Discrete Fourier Transform (DFT), denoted as $\mathcal{F}$, to map the input $\textbf{x}$ into the frequency domain, obtaining its frequency representation. The input image $\textbf{x}$ has dimensions $H \times W$ and a spatial pixel value $\textbf{x}(h,w)$ at position $(h,w)$. The Fourier transform is defined as follows. 
\begin{equation}
\mathcal{F}(\textbf{x})(a,b) = \sum_{h=0}^{H-1} \sum_{w=0}^{W-1} 
e^{-i2\pi \left(\frac{ah}{H} + \frac{bw}{W} \right)} \textbf{x}(h,w)
\end{equation}
where the output $\mathcal{F}(\textbf{x})(a,b)$ is a complex-valued frequency representation, and $(a, b)$ are spatial frequency indices representing each frequency component of the transformed image. In this complex frequency representation, let $Im$ and $Re$ denote the imaginary and real parts, respectively. The amplitude and phase of the frequency components can be expressed as:
\begin{align}
    \mathcal{A} = \left| \mathcal{F}(\textbf{x})(a,b) \right|\\
    \phi=\tan^{-1}\frac{Im(\mathcal{F}(\textbf{x})(a,b))}{Re(\mathcal{F}(\textbf{x})(a,b))}
\end{align}
Next, to extract high-level structural information from the image, we focus on capturing the high-frequency components, which represent rapid intensity variations such as sharp edges and fine details. To achieve this, we apply a high-frequency filter that removes low-frequency amplitude components. Since the phase spectrum encodes the spatial structure of the image, we preserve the complete phase information to retain structural integrity. Therefore, our high-frequency filter for amplitude masking is:
\begin{equation}
N(u,v) = \left\{
\begin{aligned}
1, & \quad if \quad d((c_u,c_v),(u,v))>r\\
0, & \quad else
\end{aligned}
\right.
,\quad \mathcal{A}_m = \mathcal{A} \odot N
\end{equation}
After obtaining the masked amplitude $\mathcal{A}_m$ and retaining all phase information $\phi$, we apply the inverse Fourier transform, denoted as $\mathcal{F}^{-1}$, to the frequency domain representation $\textbf{x}_f$. This transformation reconstructs the image back to the spatial domain, as shown below:
\begin{align}
    \textbf{x}_f = A_m * e^{j*\phi}\\
    \textbf{x} = \mathcal{F}^{-1}(\textbf{x}_f)
\end{align}

\begin{figure*}[htbp!] 
    \centering
    \includegraphics[width=\textwidth]{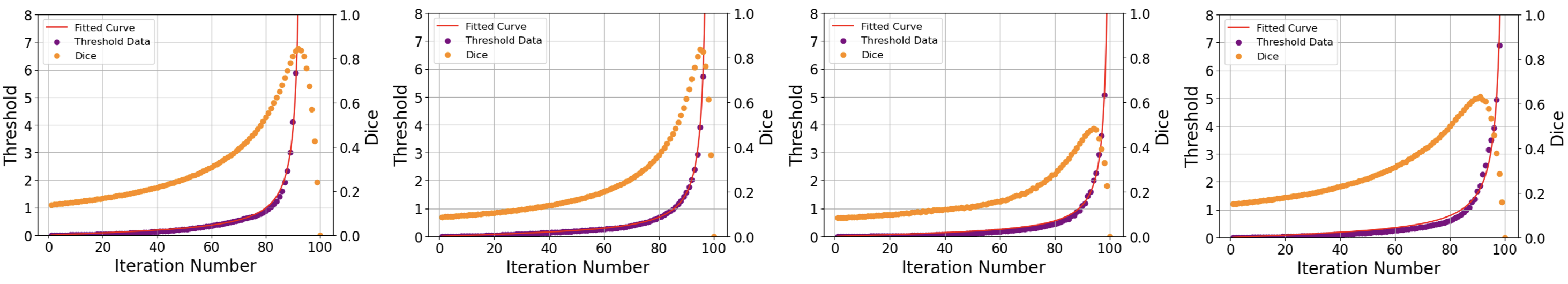} 
    \caption{
\textbf{Illustration of threshold dynamics with fixed shrinking speed of the mask.} The figures shown are plotted from four randomly chosen samples, two from the BraTS FLAIR sequence (left) and two from the BraTS T2 sequence (right). For each sample, the mask is progressively shrunk by 1\% of the brain volume per iteration until it fully disappears. 
The purple points represent the threshold required to produce the current shrunken mask based on the reconstruction error map, 
with the red curve showing the fitted function. Orange points indicate the corresponding Dice scores between the mask and the ground truth anomaly at each iteration. As the mask approaches the anomaly, the Dice score increases and peaks when the anomaly is optimally exposed. Correspondingly, the corresponding threshold rises gradually, then sharply increases near the anomaly boundary—indicating a distinct shift in reconstruction error and serving as a marker for finding the optimal stopping threshold.
    }
    \label{fig:thres}
\end{figure*}

\subsection{Subject-Specific Thresholding Strategy}
\label{sec:thres}

In this section, we discuss in detail the threshold strategy used to identify, for each image, in each iteration, which regions of the error map can be confidently identified as normal.

The original 2D approach in our previous work \citep{liang2024itermask} applies a single fixed threshold $\tau$ across all images within one sequence, and the threshold is selected using a validation set's error map, under the assumption that it reflects the model’s reconstruction error on unseen normal tissue. Thus, any error exceeding this threshold is then considered indicative of an anomaly. However, we identified two limitations that come with this approach when extending it to 3D MRI:

(1) For 3D datasets, it is usually impractical to acquire in-distribution validation data from the same clinical center with no distribution shift. 

(2) One single threshold per dataset might not be optimal for each individual subject, 
as normal tissues can exhibit subject-specific intensity distributions and characteristics.

To address these limitations, we develop a per-image iteration-stopping strategy that does not rely on a threshold derived from external healthy validation data. This enables adaptive and subject-specific anomaly detection without requiring prior knowledge of the underlying distribution of normal tissue. 
\
For each image, we observe a distinct pattern about the value of reconstruction error at which we need to apply a threshold to shrink the mask by a given percent at each iteration of our process. A gradual and smooth increase of the threshold is required to progressively shrink the mask by a small percent over normal tissue between consecutive iterations. However, once the shrinking mask reaches the anomalous region, the threshold required to shrink the mask by a small percent rises abruptly. 
This sharp change reflects the significant reconstruction error triggered by the anomaly, highlighting a clear distinction between normal and abnormal tissue. 

We extend our method to quantify and leverage  this property for testing. During the iterative testing process, at each iteration $t$ we apply a threshold $\tau(t)$ to the reconstruction error mask such that the mask is shrunk by 1\% of the brain volume per iteration, until the mask fully disappears. We show this process in Fig.~\ref{fig:thres}. As the shrinking mask approaches the anomaly, which is shown with the Dice score with yellow dots in Fig.~\ref{fig:thres}, the reconstruction error increases sharply, resulting in an abrupt change required for the threshold $\tau(t)$, as shown from the purple dots in Fig.~\ref{fig:thres}. This sudden shift enables us to identify an image-specific threshold for anomaly detection.

To find this abrupt changing point, we first estimate how the threshold changes with respect to iterations. We use a tangent parametric model to model the threshold dynamics, fitting its parameters $a$ and $b$ using least squares:
\begin{equation}
\tau(t) = a \tan(b t)
\end{equation}

This is shown by the red curve in Fig.~\ref{fig:thres}. Using the fitted function $\tau(t)$, we obtain a smooth representation of the threshold curve, that can be used to compute its derivative $\tau'(t)$. We then define the stopping threshold $\tau_{stop}$, which we use in Algorithm~\ref{algo1} to produce the final segmentation mask, as the value corresponding to the rate of change $\tau'(t)$ that first exceeds a predefined value $r$:
\begin{equation}
\text{Find } \tau_{stop}=\tau(t), \text{ such that } \tau'(t) = a b \left( \frac{1}{\cos(b t)} \right)^2 > \gamma
\end{equation}

As we are using fit function, we need to analysis the potential risk of fit-function model failure. In cases where anomalies affect the image at a global scale—such as substantial shifts in intensity distribution or large-area corruption—the dynamics of threshold variation deviate from the expected behavior. Consequently, these dynamics can no longer be reliably  modeled using a tangent-based approximation, resulting in model fitting failure.

We used $R-Squared$ coefficient of determination to measure how much of the variance in the fitting dynamics varies from the tangent curve. 
\begin{align*}
\text{SS}_{\text{res}} &= \sum_{t=1}^{80} (y_t - \tau(t))^2 \\
\text{SS}_{\text{tot}} &= \sum_{t=1}^{80} (y_t - \bar{y})^2 \\
R^2 &= 1 - \frac{\text{SS}_{\text{res}}}{\text{SS}_{\text{tot}}}
\end{align*}
Here, $y$ is the actual threshold at iteration $t$. We evaluate performance within the first 80 iterations, during which the threshold remains within a stable range. 
If the $R-Squared$ value of the fitted curve is below 0.85, we consider the model to exhibit poor fit. In such cases, we propose using discrete derivatives in place of the fitted function for the current sample.
Specifically, we compute the discrete derivative of the threshold values, apply a moving average to smooth these derivative values, and use the resulting smoothed derivative as a substitute for the first derivative of the fitted function.

\subsection{Training Process with Spatial Random Masking}
\label{sec:train}
Here, we describe the training process of our model. Our goal is to have a reconstruction model that can spatially reconstruct missing areas from the image with high-frequency guidance. 
To do this, we first synthesize spatially masked images from our training set $D_{tr}=\left\{\mathbf{x}^{i}\right\}^N_{i=1}$ which is comprised of `normal' in-distribution images. Since there is no anomaly in the training set to mask from, we generate randomly shaped masks with the help of a Gaussian distribution. Firstly, we select a random point in the brain area where $\textbf{m}$ is the 3D mask of the brain: $(\mu_x,\mu_y, \mu_z) \sim \textbf{m}$ to simulate the center of the anomaly. Then, we generate a multivariate 3D Gaussian distribution with the selected point $(\mu_x,\mu_y, \mu_z)$ as the center of the distribution of the mask to generate a random shape.
The probabilistic density function of the anisotropic 3D Gaussian distribution is 
\begin{equation}
p(x,y,z) = \frac{1}{\sqrt{(2\pi)^3 |\Sigma|}} 
\exp \left( -\frac{1}{2} (\mathbf{r} - \boldsymbol{\mu})^T \Sigma^{-1} (\mathbf{r} - \boldsymbol{\mu}) \right)
\end{equation}
Here, \( \mathbf{r} = \begin{bmatrix} x & y & z \end{bmatrix}^T \) is the position vector, and \( \boldsymbol{\mu} = \begin{bmatrix} \mu_x & \mu_y & \mu_z \end{bmatrix}^T \) denotes the mean vector (center of the distribution). The anisotropic covariance matrix \( \Sigma \) controls the spread and shape of the Gaussian distribution, where \( |\Sigma| \) is its determinant, and \( \Sigma^{-1} \) is its inverse. 
To ensure the dense region of the Gaussian distribution covers part of the brain but not the entire brain, we sample \( \Sigma \) by drawing eigenvalues $\lambda_1, \lambda_2, \lambda_3$ uniformly from [0.3,10] constructing the covariance matrix using a randomly sampled orthogonal matrix, as follows:
\begin{equation}
    \Sigma = O \,\mathrm{diag}(\lambda_1, \lambda_2, \lambda_3)\, O^T.
\end{equation}
The probability density function $p(x,y,z)$ determines the likelihood of sampling points from the distribution to serve as part of the masked area. we sample 100,000 points from the corresponding 3D multivariate normal distribution, and these points compose the random shaped mask $\mathbf{m}_r$, as illustrated in Fig.~\ref{fig:main} (b). To generate masks of varying sizes, the initial masks are upsampled by a factor of up to 4. Sampled points that fall outside the boundaries of the brain mask are discarded to ensure that the mask generated remains within the valid brain region.
 
After the mask is generated, the input image \( \textbf{x} \) is corrupted with Gaussian noise \( \epsilon \) in the masked area $\hat{\textbf{x}}$ and then fed into the model $f$, conditioned on the high-frequency image $\textbf{x}_f$ discussed in Section~\ref{sec:metfreq}, for reconstruction.
\begin{align}
    \hat{\textbf{x}} = \epsilon \odot \mathbf{m}_r + \mathbf{x} \odot (1-\mathbf{m}_r) \\
   \textbf{x}'=f(\hat{\textbf{x}}, \textbf{x}_f)
\end{align}
The network $f$ follows a UNet architecture and is trained by minimizing the $L_2$ distance loss between the reconstructed image and the original image.
\begin{equation}
    L = \lVert x'-\mathbf{x} \rVert _2
\end{equation}
Building upon \citep{liang2024itermask}, we employ a single unified model for the entire mask-shrinking process, eliminating the need to separate the first iteration.


\begin{figure}[!t]  
    \centering
    \scalebox{0.85}{ 
    \includegraphics[width=\columnwidth]{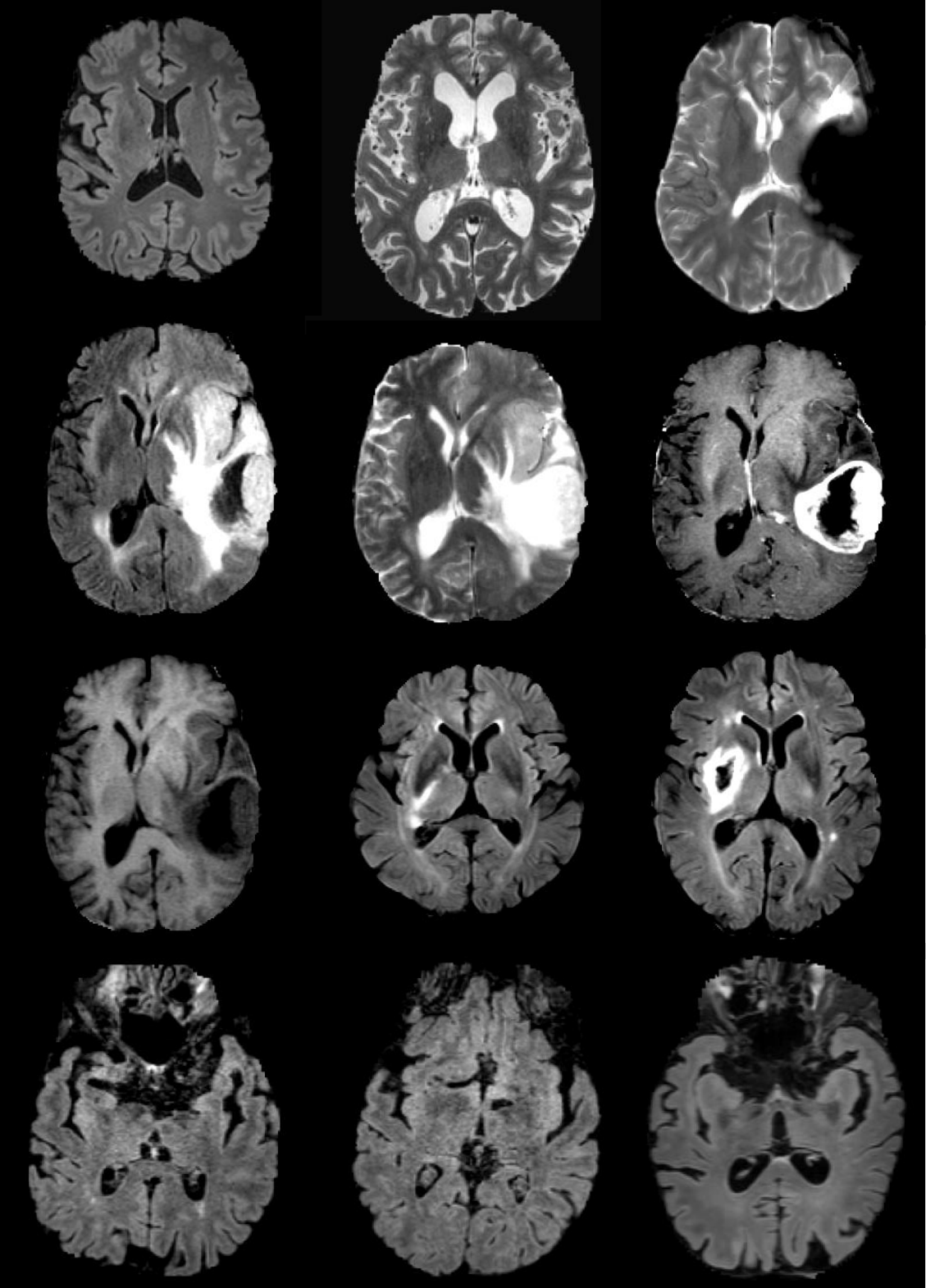}  
    }
    \caption{Illustration of the datasets used in the format: dataset name (sequence). First row, left to right, ADNI (FLAIR), OASIS (T2), Private Artifact dataset (T2); Second row, left to right, BraTS (FLAIR), BraTS (T2), BraTS (T1ce); Third row, left to right, BraTS (T1), ISLES (FLAIR) (with small lesion), ISLES (FLAIR) (with big lesion); Fourth row, three samples from ADNI (FLAIR) with problematic skull-stripping.}
    \label{fig:tumors}
\end{figure}


\begin{table*}[t!]
    \caption{Performance for detecting anomalous synthetic imaging artifacts in 3D scans. Best in bold.}
    \label{tab1}
    \centering
    \scalebox{0.8}{
\begin{tabular}{lcllccccllccc}
\hline
\multicolumn{7}{c}{\textbf{different MRI sequence (T2)}} & \multicolumn{6}{c}{\textbf{Gaussian noise}} \\ \hline
 & AUROC & AUPRC & FPR80 & FPR90 & FNR80 & \multicolumn{1}{c|}{FNR90}  & AUROC & AUPRC & FPR80 & FPR90 & FNR80 & FNR90 \\ \hline
\multicolumn{1}{l|}{AE} & 84.8 & 82.5 & 27.6 & 48.3 & 60.0 & \multicolumn{1}{c|}{84.0} & 83.6 & 83.6 & 17.4 & 69.6 & 78.3 & 78.3 \\
\multicolumn{1}{l|}{DDPM} & 78.4 & 82.0 & 48.2 & 55.2 & 66.7 & \multicolumn{1}{c|}{66.7}  & 72.0 & 71.1 & 65.2 & 65.2 & 78.3 & 78.3 \\
\multicolumn{1}{l|}{Auto DDPM}  & 82.3 & 78.4 & 17.2 & 48.2 & 80.0 & \multicolumn{1}{c|}{80.0} & 76.6 & 75.0 & 41.4 & 51.7 & 75.9 & 89.7 \\
\multicolumn{1}{l|}{Cycl. Unet} & \textbf{100.0} & \textbf{100.0} & \textbf{0.0} & \textbf{0.0} & \textbf{0.0} & \multicolumn{1}{c|}{\textbf{0.0}} & 95.7 & 96.7 & \textbf{0.0} & 21.7 & 17.4 &  17.4 \\
\multicolumn{1}{l|}{DAE (orig)} & \textbf{100.0} & \textbf{100.0} & \textbf{0.0} & \textbf{0.0} & \textbf{0.0} & \multicolumn{1}{c|}{\textbf{0.0}} & 93.8 & 90.0 & 8.7 & 8.7 & 47.8 & 87.0 \\
\multicolumn{1}{l|}{DAE (tuned)} & \textbf{100.0} & \textbf{100.0} & \textbf{0.0} & \textbf{0.0} & \textbf{0.0} & \multicolumn{1}{c|}{\textbf{0.0}} & 98.5 & 98.3 & 4.3 & 4.3 & 34.8 & 34.8 \\ 
\multicolumn{1}{l|}{IterMask3D} & \textbf{100.0} & \textbf{100.0} & \textbf{0.0} & \textbf{0.0} & \textbf{0.0} & \multicolumn{1}{c|}{\textbf{0.0}} & \textbf{100.0} & \textbf{100.0} & \textbf{0.0} & \textbf{0.0} & \textbf{0.0} & \textbf{0.0} \\ \hline

\multicolumn{7}{c}{\textbf{chunk missing (middle)}} & \multicolumn{6}{c}{\textbf{chunk mising (top)}} \\ \hline
 & AUROC & AUPRC & FPR80 & FPR90 & FNR80 & \multicolumn{1}{c|}{FNR90} & AUROC & AUPRC & FPR80 & FPR90 & FNR80 & FNR90 \\ \hline
\multicolumn{1}{l|}{AE} & 48.8 & 55.5 & 87.0 & 91.3 & 78.3 & \multicolumn{1}{c|}{82.6}  & 44.8 & 51.4 & 86.2 & 96.6 & 79.3 & 89.7 \\
\multicolumn{1}{l|}{DDPM} & 60.5 & 56.5 & 65.5 & 82.8 & 79.3 & \multicolumn{1}{c|}{89.7} & 59.3 & 60.9 & 62.1 & 86.2 & 79.3 & 79.3 \\
\multicolumn{1}{l|}{Auto DDPM} & 61.4 & 57.6 & 51.7 & 55.2 & 79.3 & \multicolumn{1}{c|}{89.7} & 69.2 & 64.8 & 37.9 & 55.2 & 75.9 & 86.2 \\
\multicolumn{1}{l|}{Cycl. Unet} & 58.4 & 68.7 & 82.6 & 95.6 & 73.9 & \multicolumn{1}{c|}{73.9} & 53.5 & 65.8 & 95.7 & 100.0 & 78.3 & 78.3 \\
\multicolumn{1}{l|}{DAE (orig)} & 70.3 & 66.4 & 39.1 & 39.1 & 70.0 & \multicolumn{1}{c|}{87.0} & 62.0 & 60.8 & 56.5 & 69.6 & 78.3 & 87.0 \\
\multicolumn{1}{l|}{DAE (tuned)} & 79.6 & 74.5 & 30.4 & 34.8 & 69.6 & \multicolumn{1}{c|}{69.6} & 65.6 & 62.5 & 52.2 & 69.6 & 78.3 & 87.0 \\
\multicolumn{1}{l|}{IterMask3D} & \textbf{95.5} & \textbf{96.7} & \textbf{0.0} & \textbf{17.4} & \textbf{13.0} & \multicolumn{1}{c|}{\textbf{13.0}} & \textbf{82.0} & \textbf{87.5} & \textbf{30.4} & \textbf{87.0} & \textbf{43.5} & \textbf{43.5} \\ \hline

\multicolumn{7}{c}{\textbf{spike noise}} & \multicolumn{6}{c}{\textbf{bias field}} \\ \hline
 & AUROC & AUPRC & FPR80 & FPR90 & FNR80 & \multicolumn{1}{c|}{FNR90} & AUROC & AUPRC & FPR80 & FPR90 & FNR80 & FNR90 \\ \hline
\multicolumn{1}{l|}{AE} & 91.3 & 93.7 & 8.7 & 56.5 & 30.4 & \multicolumn{1}{c|}{30.4} & 73.2 & 73.4 & 47.8 & 73.9 & 65.2 & 87.0 \\
\multicolumn{1}{l|}{DDPM} & 95.7 & 96.6 & 8.7 & 8.7 & 26.1 & \multicolumn{1}{c|}{26.1} & 69.6 & 60.9 & 69.6 & 69.6 & 65.2 & 65.2 \\
\multicolumn{1}{l|}{Auto DDPM} & \textbf{100.0} & \textbf{100.0} & \textbf{0.0} & \textbf{0.0} & \textbf{0.0} & \multicolumn{1}{c|}{\textbf{0.0}} & 75.6 & 81.8 & 60.9 & 65.2 & 52.2 & 52.2 \\
\multicolumn{1}{l|}{Cycl. Unet} & 95.8 & 96.7 & 0.0 & 21.7 & 17.4 & \multicolumn{1}{c|}{17.4} & 94.3 & 94.9 & 13.0 & \textbf{17.4} & 34.8 & 34.8 \\
\multicolumn{1}{l|}{DAE (orig)} & 84.5 & 77.3 & 30.4 & 30.4 & 78.3 & \multicolumn{1}{c|}{82.6} & 77.7 & 71.2 & 30.4 & 34.8 & 78.3 & 87.0 \\
\multicolumn{1}{l|}{DAE (tuned)} & 98.7 & 98.5 & 4.3 & 4.3 & 30.4 & \multicolumn{1}{c|}{30.4} & 87.5 & 86.5 & 26.1 & 26.1 & 39.1 & 82.6 \\
\multicolumn{1}{l|}{IterMask3D} & 97.2 & 97.3 & 8.3 & 12.5 & 25.0 & \multicolumn{1}{c|}{25.0} & \textbf{96.2} & \textbf{96.4} & \textbf{8.7} & \textbf{17.4} & \textbf{30.4} & \textbf{30.4} \\ \hline

\multicolumn{7}{c}{\textbf{ghosting}} & \multicolumn{6}{c}{\textbf{zipper}} \\ \hline
 & AUROC & AUPRC & FPR80 & FPR90 & FNR80 & \multicolumn{1}{c|}{FNR90} & AUROC & AUPRC & FPR80 & FPR90 & FNR80 & FNR90 \\ \hline
\multicolumn{1}{l|}{AE} & 40.3 & 45.8 & 87.0 & 100.0 & 78.3 & \multicolumn{1}{c|}{87.0} & 90.5 & 88.7 & 13.0 & 56.5 & 52.2 & 52.2 \\
\multicolumn{1}{l|}{DDPM} & 81.1 & 82.9 & 30.4 & 69.6 & 73.9 & \multicolumn{1}{c|}{73.9} & 85.3 & 86.0 & 26.1 & 39.1 & 60.9 & 60.9 \\
\multicolumn{1}{l|}{Auto DDPM} & 100.0 & 100.0 & 0.0 & 0.0 & 0.0 & \multicolumn{1}{c|}{0.0} & 94.7 & 95.0 & 8.7 & 17.4 & 39.1 & 39.1 \\
\multicolumn{1}{l|}{Cycl. Unet} & 82.0 & 86.0 & 21.7 & 78.3 & 60.9 & \multicolumn{1}{c|}{60.9} & 92.8 & 94.5 & 8.7 & 21.7 & 30.4 & 30.4 \\
\multicolumn{1}{l|}{DAE (orig)} & 91.6 & 96.0 & 4.3 & 4.3 & 0.0 & \multicolumn{1}{c|}{0.0} & 96.2 & 93.3 & 4.3 & 4.3 & 4.3 & 82.6 \\
\multicolumn{1}{l|}{DAE (tuned)} & \textbf{100.0} & \textbf{100.0} & \textbf{0.0} & \textbf{0.0} & \textbf{0.0} & \multicolumn{1}{c|}{\textbf{0.0}} & \textbf{99.6} & \textbf{99.6} & \textbf{0.0} & \textbf{0.0} & \textbf{8.7} & \textbf{8.7} \\
\multicolumn{1}{l|}{IterMask3D} & 98.0 & 99.0 & \textbf{0.0} & 8.7 & 17.4 & \multicolumn{1}{c|}{17.4} & 98.0 & 98.1 & \textbf{0.0} & 16.7 & 12.5 & 12.5 \\ \hline

\end{tabular}
}
\end{table*}

\section{Experimental Settings}
\label{sec:experiments_setting}
\label{subsec:exp_setting}
\subsection{Data}
\label{subsec:exp_setting}

We here describe the datasets used in this paper.
First, we utilize cognitively normal data from the \textbf{Open Access Series of Imaging Studies - 3 (OASIS3)} \citep{oasis3} and the \textbf{Alzheimer's Disease Neuroimaging Initiative (ADNI)} \citep{adni} datasets, as they are free from major brain pathologies and artifacts.
OASIS includes 755 cognitively normal adults and we used the T2 sequence.
The original goal of ADNI's creation was to test whether serial MRI, PET, biological markers, clinical or psychological assessment can help measure progression of mild cognitive impairment and Alzheimer's disease. Herein, from ADNI we use only the FLAIR scans from 961 cognitively-normal subjects. It is worth noting that the FLAIR sequence referred to throughout this paper specifically corresponds to the T2-weighted FLAIR (T2 FLAIR) sequence. From both datasets, 25 subjects are held-out as in-distribution test set for anomaly detection (Sec.\ref{subsec:synthetic_artifact}, \ref{subsec:real_artifact}), while the rest are used as training images.

For evaluating real artifact detection, we test our model on a private dataset acquired at our institution, consisting of 16 subjects scanned with T2-weighted imaging, wherein substantial acquisition artifacts are present. 
We also use two datasets with brain pathology, specifically \textbf{BraTS2021} with glioma cases \citep{bakas_advancing_2017,baid_rsna-asnr-miccai_2021}  and \textbf{ISLES2015} \citep{maier2017isles} 
with stroke lesion cases. We randomly selected 139 samples from BraTS2021 for testing, following the same testing sample selection as \citep{liang2024itermask, liang2023modality}. 
For ISLES2015, we used all 28 cases. 
During evaluation, if multiple lesion classes are provided, they are merged into a single class.

For experiments on 2D slices, we follow the same data pre-processing and selection process as described in \citep{liang2023modality}.
For 3D images, we first crop the image to a uniform size of $[192,192,192]$, with the brain at the center of the image. Next, we perform resampling to ensure a standardized voxel size spacing of $[1,1,1]$. Skull-stripping is then applied using Robex \citep{iglesias2011robust} to extract the brain region from the image. Finally, to exclude outliers and standardize image intensity distributions across subjects, we apply an iterative z-score normalization process constrained to the brain region. After initial z-score normalization of the brain area, we recompute the mean and variance using only pixels with intensities within three standard deviations of the current distribution to reduce effect of outliers in the statistics, and re-normalize. This step is repeated three more times.




\subsection{Evaluation Metrics}
For the task of \emph{anomaly detection}, to distinguish images with anomalies from normal images, we define a ground-truth label and an anomaly score for classification. We assign the ground-truth label to be 0 (negative) and 1 (positive) to images that are `normal' (in-distribution) and anomalous, respectively. The anomaly score is used to classify the prediction. Our method, IterMask3D, produces a final segmentation mask of any suspected anomaly in the input image. The anomaly score is then computed as the number of positive (anomalous) pixels (or voxels) in the mask. For all other methods that generate a reconstructed image, the anomaly score is calculated as the mean value of the final reconstruction error map. 
We use the above anomaly-score to evaluate anomaly detection performance with two threshold-independent classification metrics: Area Under the Receiver Operating Characteristic Curve (AUROC) and Area Under the Precision-Recall Curve (AUPRC). In addition, we provide several supplementary metrics to further evaluate performance at specific sensitivity levels. Defining as positive the `anomaly' class, and negative the `normal' class, we compute the false positive rates (FPR) and false negative rates (FNR), given that the true positive rates (TPR)are fixed at 80\% and 90\%. We refer to these metrics as FPR80, FPR90, FNR80, and FNR90 respectively.

For the task of \emph{anomaly segmentation}, we evaluate the model’s performance in delineating anomalous regions using several metrics. These include the Dice Similarity Coefficient (DSC), sensitivity, precision, and Jaccard Index, which assess how well the predicted mask overlaps the ground truth mask. DSC reflects the overall segmentation overlap, higher sensitivity indicates fewer false negatives, and higher precision corresponds to fewer false positives. 
In addition to overlap-based metrics, we also assess Average Symmetric Surface Distance (ASSD), which measures the accuracy of segmentation boundaries.
We further report the Area Under the Receiver Operating Characteristic Curve (AUROC), which evaluates the model’s ability to distinguish anomalous voxels from normal ones across all decision thresholds for each sample.
For evaluating reconstruction quality in healthy regions, we use Peak Signal-to-Noise Ratio (PSNR), which measures the pixel-level fidelity between the reconstructed image and the original input image, demonstrating the model’s ability to accurately reconstruct in-distribution tissue and thereby reduce false positives.

\subsection{Implementation Details}
For experiments on 2D slices we adopt as backbone model the network architecture from \citep{liang2024itermask}. As a 3D model we utilize a UNet with the same structure as \citep{xu2024feasibility}. The 2D model is trained with Adam optimizer with \(1e^{-4}\) learning rate for 8000 iterations with a batch size of 32. The 3D model is trained using the Adam optimizer with a learning rate of \(1e^{-3}\) for 200 epochs and a batch size of 2. The hyperparameter of the model $\gamma$ is set to 0.01 for pathology detection, and we increase it to 0.05 for 3D pathology segmentation to get a more accurate segmentation map. For 2D pathology segmentation, $\gamma$ is set to 0.01 because intensity normalization of 2D slices results to different intensity range. $r$ for frequency masking is set to 15.
For fair comparison, all baselines are implemented with the same backbone network. Experiments were implemented in PyTorch and training performed using one Nvidia V100 GPU with 24GB memory. Data augmentation is applied with random rotation at a probability of 0.1. No post-processing is used to refine the segmentation. 

\subsection{Baselines}
We include both reconstruction-based and non-reconstruction-based methods for baseline comparisons. Reconstruction-based approaches follow the `corrupt-and-reconstruct' paradigm and we include the standard autoencoder (denoted \textbf{AE}), which applies compression as corruption \citep{baur2021autoencoders}, as well as diffusion-based methods \textbf{DDPM} \citep{pinaya2022fast}, and \textbf{AutoDDPM} \cite{bercea2023mask}, which use additive noise as the corruption mechanism.
Among non-reconstruction-based methods, we evaluate a cross-sequence translation model (\textbf{Cyclic UNet}) \citep{liang2023modality}, a transformer-based method with VQ-VAE (\textbf{Transformer}) \citep{pinaya2022unsupervised} and a denoising autoencoder (\textbf{DAE}) \citep{kascenas_denoising_2022}. 

For fair comparison, we implement AE, Cyclic UNet, and DAE using 3D architectures with the same network backbone as our method. Skip connections are removed in AE. For diffusion-based methods, since no 3D implementations are available, we implement them in 2D and apply them slice-by-slice on 3D volumes; final evaluation metrics are computed in 3D. To achieve optimal performance with diffusion-based baselines, we fine-tuned the DDPM threshold used to generate the initial mask. In AutoDDPM, we improved the initial anomaly mask by switching from a fixed 2D error map proportion to a threshold-based selection, enhancing detection performance, especially on normal 3D samples. For the Transformer model, we report the results directly from the original paper \citep{pinaya2022unsupervised} because the code is not publicly available for reproducibility. To ensure a fair comparison with the original DAE method, we re-implemented its noise injection procedure. For this, we appropriately adjusted the original noise scale proportionally to the change of intensity range in our data in comparison to the original paper, to account for the using different intensity normalisation of input images (z-score) than the original paper (linear to 0-1 range). This re-implementation of the original configuration is reported as DAE(orig) in the results tables. Additionally, we tuned the noise scale by progressively increasing the noise level to identify the optimal test performance setting. The results of this tuned configuration are reported as DAE(tuned). Although this is a form of overfitting the DAE to the test data, thus giving it an unfair advantage against ours, it provides an upper bound of the DAE performance as a useful indication.

\section{Results}
\label{sec:experiments}
\subsection{Synthetic Artifact Detection on 3D Images}
\label{subsec:synthetic_artifact}

We first evaluate performance of our model for detecting artifacts that may appear as unexpected anomalies during deployment. Methods for detecting unexpected artifacts during acquisition are useful for implementing quality-assurance frameworks. 
It is often difficult to annotate precise ground truth segmentation labels for artifacts, particularly when they manifest as global patterns across the entire image. Additionally, image-level detection is typically sufficient for identifying such anomalies. Therefore we evaluate performance based on artifact detection rather than segmentation.

\begin{figure*}[!h]  
    \centering
    \scalebox{1}{ 
    \includegraphics[width=\textwidth]{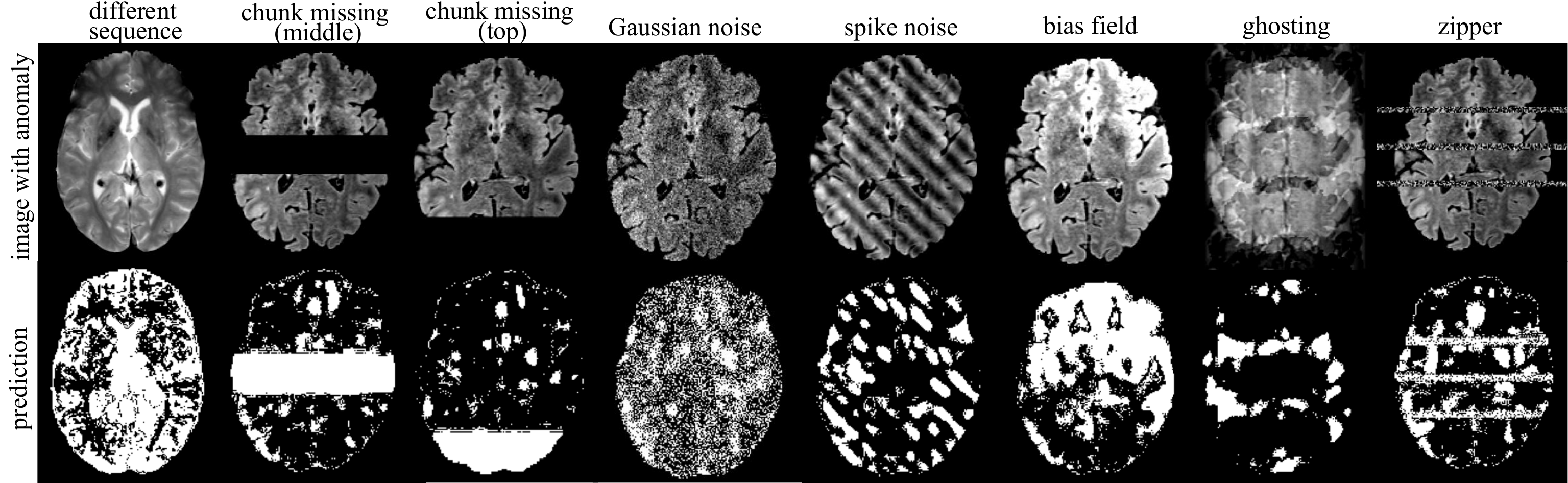}  
    }
    \caption{
    Visualization of anomaly detection results on images with synthetic anomalies using our method $IterMask3D$. The first row shows the input images with simulated anomalies, while the second row displays the corresponding detected anomaly areas.
}
    \label{fig:artifact_synth}
\end{figure*}

We begin by testing the model's ability to detect synthetic artifacts. We train models on cognitively normal data from ADNI using the FLAIR sequence, which contain no pathology or artifacts. This defines the in-distribution normal data. To evaluate anomaly detection, we first assess the model's ability to detect scans of a different sequence. Then, we simulate various types of anomalies (anomalies closer to the training distribution) in reference of \cite{graham2023latent, ravi2024efficient} by generating images with missing chunks in the middle or top regions of the brain,  adding Gaussian noise, adding spike noise, bias field, random ghosting, or zipper artifact as illustrated in Fig.~\ref{fig:artifact_synth} first row.
The simulated anomalies are added to FLAIR from the ADNI validation set, while the original, unaltered images are used as in-distribution normal control samples for performance evaluation. A detailed description of each artifact is provided below.

\begin{figure}[!t]  
    \centering
    \scalebox{1}{ 
    \includegraphics[width=1\columnwidth]{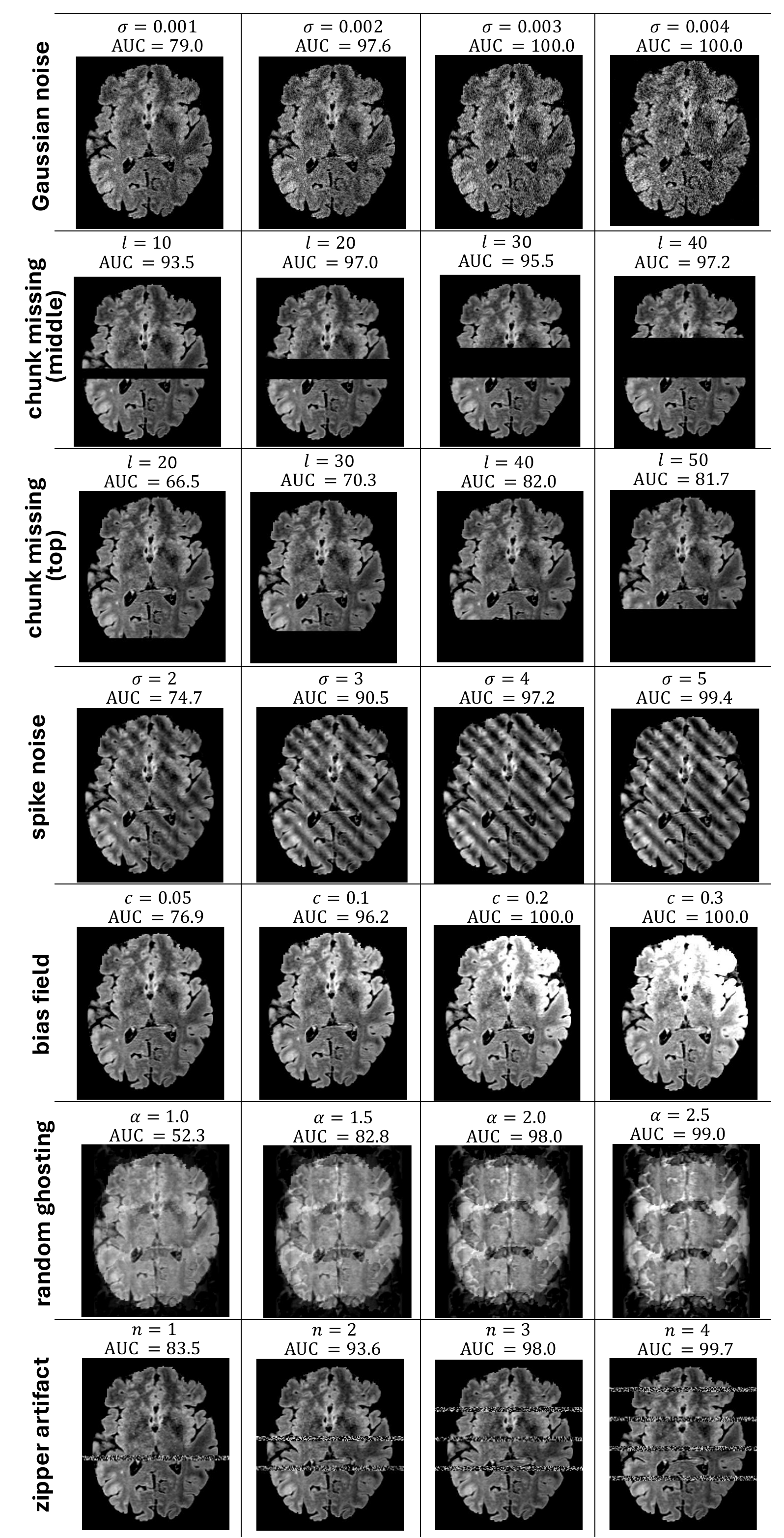}  
    }
    \caption{
    Visualization of different extents of synthetic anomaly added to the image, along with $AUC$ score of the model's anomaly detection performance.
}
    \label{fig:severity}
\end{figure}

\textbf{Different MRI sequence}:
We first evaluate our model on an MRI sequence that differs from the training modality. Specifically, the model is trained on FLAIR images, while T2 images are treated as anomalies. These are considered anomalies that are far from the training distribution.


\textbf{Missing Chunks (Top/ Middle)}: These anomalies simulate reduced field-of-view conditions, where a portion of the brain anatomy is absent from the scan. We specifically test two cases: one where the top part of the brain is missing, and another where the central middle portion is absent. We simulate missing data by applying a strip mask of width $l$ to each 2D slice of the 3D volume at the same spatial location. Specifically, we set $l = 40$ and $l = 30$ for top and middle strip-masked regions respectively for all experiments in Tab.~\ref{tab1}. We further analyze the effect of varying $l$ in Fig.~\ref{fig:severity}.


\textbf{Gaussian Noise}: Gaussian noise is usually caused by electronic disturbances within the image acquisition equipment. To simulate Gaussian noise as anomalies in an image, we add zero-mean Gaussian noise in k-space. Given $I$ as the input image:
\begin{align}
K &= \mathcal{F}(I) \\
K_{\text{complex}} &= K+K_{random}\\
I_{\mathrm{noised}} &= \mathcal{F}^{-1}(K_{\mathrm{complex}})
\end{align} 
in which $K_{\mathrm{random}}
= (\mathcal{U}\bigl(-\max(K_{\mathrm{real}} \cdot \sigma),\,\max(K_{\mathrm{real}} \cdot \sigma)\bigr) + i\,\mathcal{U}\bigl(-\max(K_{\mathrm{imag}} \cdot \sigma),\,\max(K_{\mathrm{imag}} \cdot \sigma)\bigr))$, 
\( \mathcal{U}(a, b) \) is a uniform random number between \( a \) and \( b \). $\mathcal{F}$ is Fourier transform, and $\mathcal{F}^{-1}$ is inverse Fourier transform.
\( \sigma \) is the hyper-parameter to define the amount of noise added. In Tab.\ref{tab1} included in this response, for fair comparison of all methods, we set \( \sigma \) to 0.2 for all methods, more analysis on different noise scales can be found in Fig.~\ref{fig:severity} herein.

\textbf{Spike Noise}: 
This type of noise is typically caused by small electrical discharges \cite{moratal2008k}. We simulate it by following the implementation in TorchIO \cite{perez2021torchio}, which randomly creates $n$ abnormal points in k-space by increasing the intensity of these points.
\begin{align}
K &= \mathcal{F}(I) \\
K_{\mathrm{spike}}
&= K
+ \sum_{i=1}^{n}
    \delta\bigl(k - k_i\bigr)\,\sigma\,\max\!\bigl(\lvert K\rvert\bigr) \\
I_{\mathrm{noised}} &= \mathcal{F^{-1}}(K_{\mathrm{spike}})
\end{align} 
In the above function, $\delta$ function picks out one specific point $k_i$ in k-space and $\sigma$ controls the amount of noise added. For a fair comparison of all methods in our table, we only set one abnormal point in the k-space ($n=1$), and we set the spike point to $k=[0.4,0.4,0.4]$ and $\sigma=0.2$ for all methods. We evaluate the model's performance under varying levels of anomaly by adjusting the value of $\sigma$. The corresponding results are presented in Figure~\ref{fig:severity} herein. 

\begin{table*}[t!]
    \caption{Detection of real-world imaging artifacts in 3D scans. Best in bold.}
    \label{tab:artifact2}
    \centering
    \scalebox{0.8}{ 
\begin{tabular}{l|llllll|llllll}
\hline
\multicolumn{7}{c}{\textbf{Magnetic Susceptibility Artifact (T2)}} & \multicolumn{6}{c}{\textbf{Failed skull-stripping (FLAIR)}} \\ \hline
 & \multicolumn{1}{c}{AUROC} & AUPRC & \multicolumn{1}{c}{FPR80} & \multicolumn{1}{c}{FPR90} & \multicolumn{1}{c}{FNR80} & FNR90 &{AUROC} & AUPRC & \multicolumn{1}{c}{FPR80} & \multicolumn{1}{c}{FPR90} & \multicolumn{1}{c}{FNR80} & FNR90\\ \hline
AE & 41.9 & 38.1 & 76.0 & 92.0 & 80.0 & 80.0  & 24.1 & 35.7 & 100.0 & 100.0 & 80.0 & 90.0 \\
DDPM & 45.8 & 40.4 & 63.3 & 100.0 & 80.0 & 86.7 & 22.0 & 37.9 & 91.3 & 100.0 & 80.0 & 90.0 \\
Auto DDPM &  55.3& 45.6 & 66.7 &66.7  &73.3  &  86.7 & 45.7 & 48.4 & 82.6 & 87.0 & 80.0 & 90.0 \\
Cycl. Unet & 83.2 & 75.8 & 20.0 & 52.0 & 80.0 & 80.0 & 70.9 & 72.1 & 52.2 & 65.2 & 75.0 & 75.0 \\
DAE (orig) & 98.0 & 96.4 & 3.3 & 10.0 & 26.7 & 26.7 & 88.0 & 81.5 & 13.0 & 34.8 & 55.0 & 55.0 \\
DAE (tuned) & 99.2 & \textbf{99.6} & \textbf{0.0} & \textbf{0.0} & \textbf{6.7} & \textbf{6.7} & 87.5 & 86.1 & \textbf{17.4} & \textbf{17.4} & 60.0 & 60.0 \\
IterMask3D & \textbf{99.7} & \textbf{99.6} & \textbf{0.0} & \textbf{0.0} & \textbf{6.7} & \textbf{6.7} & \textbf{90.0} & \textbf{91.9} & \textbf{17.4} & 21.7 & \textbf{35.0} & \textbf{35.0} \\ \hline
\end{tabular}
    }
\end{table*}

\textbf{Bias Field}: 
Bias fields, appearing as low-frequency variations in signal intensity,  are usually caused by inconsistencies in radio-frequency coil uniformity, magnetic field inhomogeneities, or the patient's anatomy \cite{sled2002nonparametric}. We followed TorchIO's implementation, which models the bias field as a linear combination of polynomial basis functions \cite{van1999automated}: 
\[
I_{\text{biased}}(x,y,z) = I(x,y,z) \cdot \exp \left( \sum_{i=0}^{3} \sum_{j=0}^{3-i} \sum_{k=0}^{3-(i+j)} c_{ijk} \cdot x^i \cdot y^j \cdot z^k \right)
\]
In all experiments, we ensure a fair and easy comparison between methods by setting $c_{ijk}$ to the same value across all methods. For Tab.~\ref{tab1}, we use $c_{ijk}=0.1$. We further analyze the effect of varying this shared value in Fig.~\ref{fig:severity}.

\textbf{Random Ghosting}: 
Ghosting artifacts usually appear as repeating patterns along the phase-encoding direction, resulting from regular fluctuations in the position of anatomical structures. In clinical MRI, these artifacts are often associated with movement of the patient, such as due to breathing or heartbeat, or sometimes physiological processes such as blood pulsation. They can be simulated by periodically removing specific planes from k-space.  We followed the implementation of TorchIO  \cite{perez2021torchio}. 
\begin{align}
K &= \mathcal{F}(I)\\
K_{\mathrm{ghost}}
&=
\begin{cases}
(1 - \alpha)\,K, & \text{for planes at every } n_{th},\\[6pt]
K, & \text{for other planes}.
\end{cases}\\
I_{\mathrm{noised}} &= \mathcal{F}^{-1}(K_{\mathrm{ghost}})
\end{align}
For a fair comparison, we set the hyperparameters $n$ and $\alpha$ to the same values across experiments with all methods in Tab.~\ref{tab1}. Further analysis on the impact of anomaly severity is presented in Fig.~\ref{fig:severity}, where we vary the hyperparameter $\alpha$.

\textbf{Zipper}:  Zipper artifact appears as bands of high-frequency noise running from one side of the image to the other, regularly spaced across it. It can be caused by radio-frequency signal disruptions. It can be simulated by changing the intensity value of strips of the original image as follows:
\begin{align}
I'(x, y, z) &= M(x, y) \cdot I_{\text{noise}}(x, y, z) + (1 - M(x, y)) \cdot I(x, y, z) \\
M(x, y) &= \sum_{i=1}^{n} \mathbf{1}\left[ y_{i,\text{start}} \leq y < y_{i,\text{start}} + h_i \right]
\end{align}
where $M$ is the mask for creating zipper, $I_{noise}$'s implementation follows \cite{ravi2024efficient}. $n$ refers to the number of strip added. $h$ refers to the hight of each stripe. For fair comparison, we set $h$ to 5 for all the experiments in Tab.~\ref{tab1} and Fig.~\ref{fig:severity}. And analyzed the performance by changing $n$ in Fig.~\ref{fig:severity} to evaluate the sensitivity of our model to this artifact.

Results of the above synthetic artifacts are shown in Tab.~\ref{tab1}.
When testing on MRI sequences significantly different from the training distribution or introducing global anomalies such as Gaussian noise, most reported methods effectively distinguish these anomalies from in-distribution samples. However, baseline methods typically exhibit reduced performance when faced with large missing brain regions, particularly in cases where chunks from the top or middle areas of the brain are absent. Reconstruction-based approaches generally corrupt the input images and attempt to restore the missing information, but these methods are usually trained to reconstruct only partially corrupted regions, relying heavily on surrounding contextual information. Consequently, when a substantial portion of the brain is missing—visually resembling the background (e.g. air)—these methods often fail to reconstruct meaningful structures, resulting in low reconstruction errors and undetected anomalies.

On the other hand, when the missing areas are small and lead to structural discontinuities, such as spike noise or ghosting, baseline methods with strong generative capabilities, such as diffusion-based models, can successfully reconstruct these anomalies using nearby information, thereby enabling their detection.

In contrast, our proposed method incorporates high-frequency structural information as additional guidance. This high-frequency component captures global structural patterns across the entire image, enhancing sensitivity to structural inconsistencies. As a result, our method achieves robust anomaly detection performance, effectively identifying anomalies characterized both by large missing anatomical regions and small localized discontinuities.


To further evaluate the tolerance level of our algorithm to different type of artifacts, we test the performance on different levels of anomaly added, as shown in Fig.~\ref{fig:severity}. Specifically, we adjusted the degree of synthetic anomalies introduced to examine how anomaly detection performance varies with the severity of the anomalies in terms of auc score.

\begin{figure}[!t] 
    \centering
    \scalebox{1}{ 
    \includegraphics[width=\columnwidth]{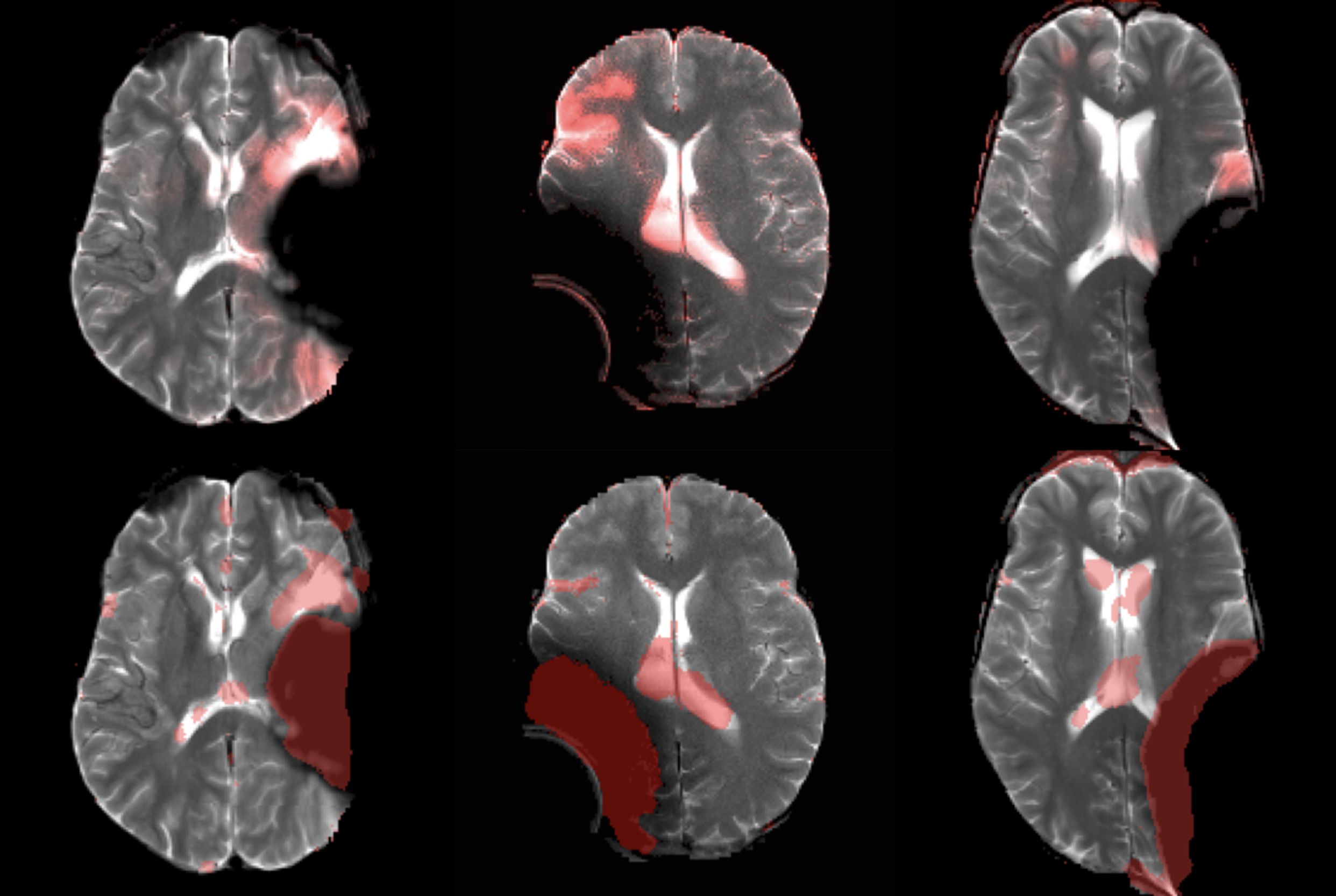}  
    }
    \caption{Visualization of artifact detection results on our private dataset using DAE \citep{kascenas_denoising_2022} (top row) and our method $IterMask3D$ (bottom row). The final anomaly map is overlaid in red on the original image. DAE is shown to detect the hyper-intense area (mostly in normal brain area) as anomaly, while our method better detects the actual anomalous artifact.}
    \label{fig:artifact}
\end{figure}

\subsection{Real-world Artifact Detection on 3D Images}
\label{subsec:real_artifact}

In this section, we evaluate our model's performance on two types of real-world imaging artifacts. The first database is scans with metallic susceptibility artifacts, which are reasonably common in MRI scans of the brain due to implanted medical devices such as programmable VP-shunt valves or cochlear implant devices. This database consists of T2 MRI sequence from our collaborative hospital. Examples are shown in Fig.~\ref{fig:tumors} (first row, third image) and Fig.~\ref{fig:artifact}. Typically, these artifacts feature a missing brain region covered by hypo-intense signal in T2, accompanied by hyper-intense signal spikes at the boundary of the affected area. 

The second database includes scans of FLAIR sequence with artifacts due to failures of skull-stripping, where skull-stripping applied during preprocessing has been suboptimal and has left residual tissues. This is common in real-world pipelines for MRI analysis, as skull-stripping methods may partially fail in various cases, affecting downstream analysis. For this purpose, we created a dataset of images with problematic skull-stripping. We processed FLAIR images of the ADNI database by applying three distinct skull-stripping algorithms to each image: HD-BET \cite{sled2002nonparametric}, Deepbet \cite{fisch2024deepbet}, and SynthStrip \cite{hoopes2022synthstrip}. We compute the pairwise Dice similarity coefficient between the 3 new brain-masks and the initial brain masks obtained during the data preprocessing stage. We then average these values. Low average Dice score indicates that a particular mask for this subject deviates notably from the others, suggesting a likely skull-stripping failure. This approach allows us to systematically identify and study failure cases based on inter-method disagreement. We then collected the 20 cases with the most severe skull-stripping failure (as per lowest Dice-score) and the associated brain-masks. Examples are shown in 4th row of Fig.~\ref{fig:tumors}. These collected images are real-world cases where different skull-stripping methods have partially failed.

All models were trained exclusively on cognitively normal cases from the OASIS dataset in the T2 sequence and ADNI dataset in FLAIR sequence. For testing purposes, we combined our private artifact dataset with the held-out T2 test samples from OASIS and ADNI without artifacts and evaluated performance using the following metrics: AUROC, AUPRC, FPR80, FPR90, FNR80, and FNR90. All methods tested use the mean value of the error map as anomaly score during evaluation.

\begin{figure*}[!t]  
    \centering
    \scalebox{1.9}{ 
    \includegraphics[width=1\columnwidth]{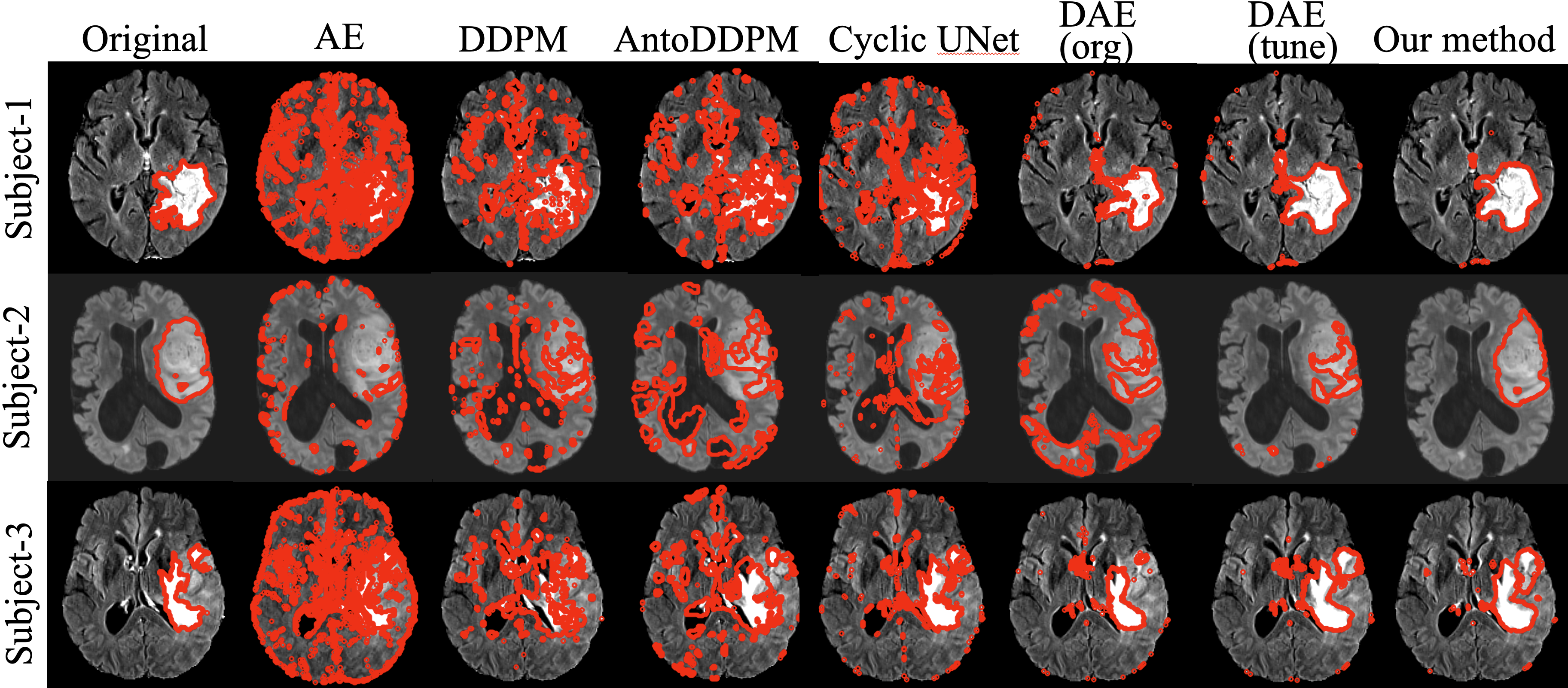}  
    }
    \caption{
    Anomaly segmentation results on 3D FLAIR inputs from BraTS using $IterMask3D$. Red contours indicate the boundaries of the predicted segmentation masks overlaid on the original images. 
}
    \label{fig:3DVIS}
\end{figure*}
\subsection{Pathology Segmentation on 3D Scans}
\label{subsec:3d_patho}

The quantitative results, shown in Tab.~\ref{tab:artifact2}, demonstrate promising performance for our method and comparable with the DAE baseline. 
However, upon visual inspection of anomaly maps (shown in Fig.\ref{fig:artifact}) for metallic susceptibility artifacts, we observe substantial qualitative differences between the two methods. As illustrated in the top row of Fig.\ref{fig:artifact}, the DAE method primarily detects hyper-intense regions near the anomaly, also including parts of normal brain tissue, but fails entirely to identify the main artifact region that appears hypo-intense, as structural discontinuity of the brain. This limitation arises because DAE has to assume a prior about the appearance of anomalies, reflected in the use of hyper-intense noise for its training; hence, it predominantly targets and removes high-intensity signals during testing, missing artifacts with low intensity value.
In comparison, as shown in the bottom row of Fig.~\ref{fig:artifact}, our method makes no strong prior assumptions about the appearance of anomalies and can identify both hypo-intense and hyper-intense artifacts, including areas with signal loss and bright boundary spikes. This allows our method to detect structural discontinuities more effectively than DAE.


\begin{table*}[h]
\centering
\caption{Performance of methods for pathology segmentation on 2D slices across all MRI sequences in BraTS. Best in bold.}
\label{tab3}
\scalebox{0.85}{
\begin{tabular}{l|cccc|cccc|cccc|cccc}
\hline
sequence & \multicolumn{4}{c|}{\textbf{FLAIR}} & \multicolumn{4}{c|}{\textbf{T1CE}} & \multicolumn{4}{c|}{\textbf{T2}} & \multicolumn{4}{c}{\textbf{T1}} \\ \hline
Metrics & DSC & Sens & Prec & SSIM & DSC & Sens & Prec & SSIM & DSC & Sens & Prec & SSIM & DSC & Sens & Prec & SSIM \\ \hline
AE & 33.4 & 54.0 & 27.0 & 27.4 & 32.3 & 50.1 & 26.4 & 33.8 & 30.2 & 62.8 & 21.1 & 33.1 & 28.5 & \textbf{94.9} & 17.4 & 28.8 \\
DDPM & 60.7 & 57.8 & 69.8 & 34.6 & 37.9 & 35.2 & 46.2 & 26.4 & 36.4 & 33.6 & 44.2 & 38.6 & 29.4 & 32.1 & 31.5 & 20.8 \\
AutoDDPM & 55.5 & 57.5 & 58.7 & 22.7 & 36.9 & 58.9 & 28.9 & 38.6 & 29.7 & 53.0 & 22.3 & 39.3 & 33.5 & 61.1 & 24.7 & 34.9 \\
Cycl.UNet & 65.0 & 63.4 & 73.9 & 24.5 & 42.6 & 47.5 & 42.9 & 29.8 & 49.5 & 48.8 & 53.4 & 30.0  & 37.0 & 45.8 & 35.2 & 25.8 \\
DAE\_pos & 79.7 & 79.1 & \textbf{84.5} & 28.0 & 36.7 & 42.0 & 36.2 & 30.0 & 69.6 & 68.1 & \textbf{75.3} & 33.7 & 29.5 & 61.2 & 20.5 & 28.3 \\
DAE\_neg & 28.5 & \textbf{94.9} & 17.3 & 18.6 & 34.7 & 37.9 & 39.3 & 32.3 & 28.5 & \textbf{94.9} & 17.4 & 32.9 & 47.9 & 53.7 & 50.9 & 25.4 \\
DAE\_full & 73.7 & 72.9 & 80.5 & 27.0 & 46.3 & 47.9 & 51.5 & 34.0 & 60.4 & 58.5 & 69.1 & 33.9 & 44.5 & 48.0 & 47.5 & 27.8 \\
$\rm{IterMask^2}$ & \textbf{80.2} & 81.3 & 83.3 & \textbf{38.0} & \textbf{61.7} & \textbf{59.1} & \textbf{70.9} & \textbf{51.6} & \textbf{71.2} & 74.4 & 72.9 & \textbf{46.2} & \textbf{58.5} & 56.6 & \textbf{67.6} & \textbf{45.2} \\ \hline
\end{tabular}
}
\end{table*}

\subsection{Pathology Segmentation on 2D Slices of Scans} 
\label{subsec:2d_patho}

In this section, we evaluate the performance of pathology segmentation on 2D MRI slices. We follow standard practices in the field \citep{pinaya2022fast,bercea2023mask} to ensure a fair comparison, where the model is trained on normal 2D slices (i.e., healthy in this experiment) extracted from 3D scans of training subjects and tested on 2D slices with pathologies extracted from 3D scans of test subjects from the same database. Anomaly segmentation performance was evaluated by how well the model is able to segment the previously unseen pathological regions. The metrics used in this experiment are Dice coefficient (DSC), sensitivity, precision, and PSNR. 

We test on all sequences of BraTS to show the model is robust to different sequences with pathologies of different intensities. Results are shown in Tab.~ \ref{tab3}.
The results show that our model performs consistently well across different sequences under different evaluation metrics, demonstrating promising generalizability. 

Here, we also conduct a detailed analysis of the baseline model DAE \citep{kascenas_denoising_2022} on the BraTS sequences. This model is trained to remove added noise and, during testing, treats anomalies as if they were noise to be removed. The original DAE adds hyper-intense noise — clipped to positive values ($[0,+\infty]$ row in the table), and shows strong performance on FLAIR (Dice: 0.797) and T2 (Dice: 0.696), but performs poorly on T1 (Dice: 0.295) and T1ce (Dice: 0.367). As shown in Fig.~\ref{fig:tumors}, tumors in the BraTS dataset appear hyper-intense in FLAIR and T2, dark in T1, and contain both hyper-intense and dark regions in T1ce. For analysis, we change the noise to the negative range ($[-\infty,0]$ as in table). Then, performance on T1 improves to 0.479. If the noise includes both positive and negative values ($[-\infty,+\infty]$), performance on T1ce improves to 0.463. These results suggest that DAE requires prior knowledge about the expected anomaly appearance to perform well, which is usually unavailable when the types of anomalies to expect is unknown prior to deployment. In comparison, our method consistently achieves better or comparable performance to DAE without relying on intensity priors, demonstrating its effectiveness.

\begin{table}[t!]
\caption{Performance for 3D pathology segmentation in BraTS and ISLES datasets. Best in bold.}
\label{tab:3dpathology}
\centering
\scalebox{0.85}{
\begin{tabular}{l@{\hskip 6pt}c@{\hskip 5pt}c@{\hskip 5pt}c@{\hskip 3pt}c@{\hskip 3pt}c@{\hskip 3pt}c@{\hskip 3pt}c}
\hline
\multicolumn{8}{c}{\textbf{BraTS FLAIR}} \\ \hline
& DSC & AUC & Sensitivity & Precision & Jaccard & PSNR & ASSD \\ \hline
\multicolumn{1}{l|}{AE} & 16.0 & 92.6 & 42.8 & 10.7 & 8.9 & 29.1 & 17.4 \\
\multicolumn{1}{l|}{DDPM} & 28.5 & 95.7 & 55.4 & 23.5 & 18.9 & 31.3 & 13.3 \\
\multicolumn{1}{l|}{Auto DDPM} & 21.5 &  80.7  &27.9  & 21.6 & 12.7 &26.8 & 16.9\\
\multicolumn{1}{l|}{Cycl. UNet} & 40.0 & 96.9 & 45.9 & 39.5 & 28.3 & 13.5 & 26.3 \\
\multicolumn{1}{l|}{Transformer} & 61.7 & - & - & - & - & - & - \\
\multicolumn{1}{l|}{DAE (orig)} & 60.0 & 98.7 & 63.8 & 60.9 & 46.2 & 33.6 & 9.2 \\
\multicolumn{1}{l|}{DAE (tuned)} & 66.7 & 99.1 & 71.1 & 68.7 & 52.5 & \textbf{34.5} & 6.2 \\
\multicolumn{1}{l|}{IterMask3D} & \textbf{69.7} & \textbf{99.2} & \textbf{72.7} & \textbf{71.1} & \textbf{56.7} & 32.2 & \textbf{5.1} \\ \hline

\multicolumn{8}{c}{\textbf{BraTS T2}} \\ \hline
& DSC & AUC & Sensitivity & Precision & Jaccard & PSNR & ASSD \\ \hline
\multicolumn{1}{l|}{AE} & 12.4 & 90.5 & 66.1 & 7.1 & 6.8 & 17.7 & 30.7 \\
\multicolumn{1}{l|}{DDPM} & 18.9 & 92.8 & 42.6 & 14.1 & 11.6 & 35.6 & 16.2 \\
\multicolumn{1}{l|}{Auto DDPM} & 19.4  & 77.8 &29.4  &16.2& 11.1 &  27.2 & 18.13\\
\multicolumn{1}{l|}{Cycl. UNet} & 20.2 & 93.6 & 47.3 & 13.7 & 11.8 & 26.5 & 17.7\\
\multicolumn{1}{l|}{DAE (orig)} & 37.9 & 97.1 & 48.8 & 35.8 & 24.2 & \textbf{37.5} & 14.5 \\
\multicolumn{1}{l|}{DAE (tune)} & 53.6 & \textbf{97.9} & \textbf{57.0} & \textbf{56.4} & 38.3 & 1.0 & \textbf{9.5} \\
\multicolumn{1}{l|}{IterMask3D} & \textbf{53.8} & 97.7 & 55.4 & 55.9 & \textbf{39.6} & 36.7 & \textbf{9.5} \\ \hline

\multicolumn{8}{c}{\textbf{ISLES FLAIR}} \\ \hline
& DSC & AUC & Sensitivity & Precision & Jaccard & PSNR & ASSD \\ \hline
\multicolumn{1}{l|}{AE} & 7.4 & 93.3 & 53.3 & 4.8 & 4.1 & 26.9 & 24.5 \\
\multicolumn{1}{l|}{DDPM} & 12.1 & 95.2 & \textbf{58.5} & 9.4 & 7.9 & 30.4 & 21.7 \\
\multicolumn{1}{l|}{Auto DDPM} & 6.94 & 82.6 & 45.7 & 5.1& 4.1 & 24.15 & 25.17 \\
\multicolumn{1}{l|}{Cycl. UNet} & 15.1 & 94.9 & 32.8 & 12.8 & 9.7 &  24.1 & 24.3 \\
\multicolumn{1}{l|}{DAE (orig)} & 30.3 & 97.6 & 42.3 & 30.3 & 21.5 & 31.1 & 22.2 \\
\multicolumn{1}{l|}{DAE (tune)} & \textbf{33.0} & \textbf{97.8} & 42.4 & \textbf{33.2} & \textbf{24.1} & \textbf{32.0} & 21.7 \\
\multicolumn{1}{l|}{IterMask3D} & 31.4 & 97.2 & 40.3 & 30.4 & 23.0 & 29.3 & \textbf{20.3} \\ \hline
\end{tabular}
}
\end{table}

In this section, we test the performance on whole 3D scans and aim to segment pathologies as anomalies. The models are trained using cognitively normal subjects from the ADNI and OASIS datasets as described in section \ref{subsec:exp_setting}. We test the models on FLAIR and T2 sequences of BraTS, as well as FLAIR of the ISLES2015 dataset. 

We assess segmentation performance using several metrics: Dice coefficient (DSC), AUROC, sensitivity, precision, Jaccard index, and Average Symmetric Surface Distance (ASSD). Additionally, we compute the Peak Signal-to-Noise Ratio (PSNR) within the healthy brain regions (excluding pathology) to evaluate the model’s ability to accurately reconstruct in-distribution, normal tissue.

Results are presented in Tab.~\ref{tab:3dpathology} and visualization results are presented in Fig.~\ref{fig:3DVIS}. Our method outperforms all baseline approaches that do not rely on prior knowledge of anomaly appearance during training. In particular, on the FLAIR sequence of the BraTS dataset, our model shows clear improvements over AE, DDPM, Auto-DDPM, Cyclic UNet, and Transformer-based method. For the DAE baseline, the original implementation uses noise with 0.2 standard deviation, and due to the altered range of intensities in our data from z-score normalization, we rescale it to the corresponding 1.2 standard deviations. For an even stronger comparison, we further tune DAE's noise level to 3 standard deviations, value found via configuration experiments aiming to maximize its performance on the test set. Although this gives it an edge over our method, as it is a form of overfitting the test set, it optimally matches the DAE's intensity prior to these anomalies and gives an indication about its optimal potential. Both DAE configurations are reported in Tab.~\ref{tab:3dpathology} as DAE (1.2std) and DAE (3std). Notably, our method achieves comparable performance to the over-tuned DAE(3std), despite not incorporating any intensity-based prior knowledge.

\begin{table}[h!]
    \caption{Ablation study for observing the effect of spatial and frequency masking in the 2D (top) and 3D (bottom) versions of our model. We also show performance of our 2D model (IterMask$^2$) in 3D images from BRATS (bottom) for straightforward comparison with IterMask3D, showing the gains from 3D modeling.}
    \label{tab:ablation1}
    \centering
    \scalebox{0.85}{
\begin{tabular}{lccc}
\hline
 & DSC & Sensitivity & Precision \\ \hline
\multicolumn{4}{l}{\textbf{2D BraTS FLAIR (2D Metric)}} \\ \hline
 \multicolumn{1}{l|}{$\rm{IterMask^2}$} & 80.2 & 81.3 & 83.3 \\ 
\multicolumn{1}{l|}{frequency  masking only} & 75.1 & 73.8 & 79.6 \\
\multicolumn{1}{l|}{spatial masking only} & 64.2 & 69.4 & 63.6 \\ \hline
\multicolumn{4}{l}{\textbf{3D BraTS FLAIR (3D Metric)}} \\ \hline
\multicolumn{1}{l|}{$\rm{IterMask^2}$} & 49.7 & 46.8 & 77.8 \\ \hdashline
\multicolumn{1}{l|}{Itermask3D} & 69.7 & 72.7 & 71.1 \\
\multicolumn{1}{l|}{frequency  masking only} & 60.8 & 65.9 & 64.8 \\
\multicolumn{1}{l|}{spatial masking only} & 40.4 & 43.7 & 42.2 \\ \hline
\end{tabular}
}
\end{table}

\subsection{Ablation Study}
Finally, we conduct ablation studies to evaluate the effectiveness of 
spatial masking and frequency masking, 3D model compared to 2D, and new subject-specific thresholding strategy. 

When performing ablation study on \textit{spatial and frequency masking}, the ablation experiments are repeated using both 2D and 3D models for the task of pathology segmentation, as described in the previous two sections, using the FLAIR sequence of the BraTS dataset. Results are shown in Tab.~\ref{tab:ablation1}.
To evaluate the role of spatial masking, we remove the spatial mask refinement process and input only the high-frequency structural information to the network. The anomaly score is then computed from a single-step reconstruction, denoted as \textit{frequency masking only} in Tab.~\ref{tab:ablation1}.
Next, to evaluate the impact of frequency masking, we remove the high-frequency-based auxiliary input and train the model using only spatially masked images as input. The full iterative reconstruction process is retained, but without high-frequency structural guidance. Results for this variant are reported as \textit{spatial masking only} in Tab.~\ref{tab:ablation1}. Results in all settings demonstrate the positive contribution of both components.

To evaluate the performance of \textit{the 3D model compared to 2D},
we retrained and tested the 2D model using the same training and testing splits, preprocessing pipeline, and experimental settings as the 3D model. To ensure consistency in evaluation, we applied the same validation-set thresholding method previously used in our 2D experiments, since subject-specific thresholding is not applicable when generating 3D predictions slice by slice. We then tested the model on the task of segmentation of anomalies on 3D FLAIR imaging from BRATS, where tumors are considered anomalies. We evaluated the model on all slices of the test images and computed the corresponding 3D evaluation metrics. The results of this additional experiment are included in Tab.~\ref{tab:ablation1} on \textit{$\rm{IterMask^2}$} row of the 3D BraTS FLAIR section. The results show that a 3D model is clearly superior to a 2D model. 

To test the performance of the newly proposed subject-specific thresholding strategy, we conducted two ablation studies with $IterMask^2$ and $IterMask3D$ respectively. 
For $IterMask^2$, we compare two thresholding strategies on the task of 2D anomaly segmentation on the BraTS dataset on all four sequences: FLAIR, T1CE, T2, and T1, as shown in Tab.\ref{tab:ablation2}. In the table, \textit{fixed} refers to the fixed threshold defined via cross-validation set on normal data, and \textit{adapt} refers to the new \emph{proposed} subject-specific thresholding strategy.  As can be seen from the results, 
the enhanced method achieves better or comparable performance, demonstrating the effectiveness of the proposed thresholding approach.

For $IterMask3D$, we compare two thresholding strategies on the task of 3D real-world anomaly detection. The anomalies include \textit{metallic susceptibility artifacts in T2 sequence} collected from our collaborative hospital and \textit{failed skull-stripping in FLAIR sequence} cases from the ADNI dataset. The training dataset used for the former experiment is OASIS dataset T2 sequence, whereas for the latter we used ADNI dataset FLAIR sequence. The results are presented in Tab.~\ref{tab:ablation3}. Here, same as in Tab.\ref{tab:ablation2}, \textit{fixed} uses a cross-validated threshold on normal data, while \textit{adapt} applies the \emph{proposed} subject-specific strategy. From the results, we observe that the subject-specific thresholding improves performance on both datasets. This suggests that when there is a domain shift between the training and testing datasets, a threshold derived from using the training ata may not generalize well to an unseen test domain. Our adaptive subject-specific threshold strategy alleviates this.

\begin{table}[t!]
    \caption{Ablation study of thresholding strategy with $IterMask^2$ on 2D anomaly segmentation.}
    \label{tab:ablation2}
    \centering
    \scalebox{0.85}{
\begin{tabular}{lccc}
\hline
& DSC & Sensitivity & Precision \\ \hline
\multicolumn{4}{l}{\textbf{ BraTS FLAIR}} \\ \hline
\multicolumn{1}{l|}{$\rm{IterMask^2}$ - fixed} & 80.2 & 81.3 & 83.3 \\
\multicolumn{1}{l|}{$\rm{IterMask^2}$ - adapt} & 80.4 & 80.6 & 83.1 \\ \hline
\multicolumn{4}{l}{\textbf{ BraTS T1CE}} \\ \hline
\multicolumn{1}{l|}{$\rm{IterMask^2}$ - fixed} & 61.7 & 59.1 & 70.9 \\
\multicolumn{1}{l|}{$\rm{IterMask^2}$ - adapt} & 64.3 & 62.6 & 69.8 \\ \hline
\multicolumn{4}{l}{\textbf{ BraTS T2}} \\ \hline
\multicolumn{1}{l|}{$\rm{IterMask^2}$ - fixed} & 71.2 & 74.4 & 72.9 \\
\multicolumn{1}{l|}{$\rm{IterMask^2}$ - adapt} & 71.4 & 72.2 & 70.2 \\ \hline
\multicolumn{4}{l}{\textbf{ BraTS T1}} \\ \hline
\multicolumn{1}{l|}{$\rm{IterMask^2}$ - fixed} & 58.5 & 56.6 & 67.6 \\
\multicolumn{1}{l|}{$\rm{IterMask^2}$ - adapt} & 59.6 & 57.7 & 67.6 \\ \hline
\end{tabular}
}
\end{table}

\begin{table}[t!]
    \caption{Ablation study of thresholding strategy with $IterMask3D$ on 3D real-world anomaly detection.}
    \label{tab:ablation3}
    \centering
    \scalebox{0.85}{
\begin{tabular}{lcc}
\hline
 & AUROC & AUPRC  \\ \hline
\multicolumn{3}{l}{\textbf{Metallic Susceptibility Artifact}} \\ \hline
\multicolumn{1}{l|}{Itermask3D - fixed} & 89.3 & 91.8 \\
\multicolumn{1}{l|}{Itermask3D - adapt} & 99.7 & 99.6  \\ \hline
\multicolumn{3}{l}{\textbf{Failed skull-stripping}} \\ \hline
\multicolumn{1}{l|}{Itermask3D - fixed} & 72.0 & 76.6 \\
\multicolumn{1}{l|}{Itermask3D - adapt} & 86.1  & 87.5 \\ \hline
\end{tabular}
}
\end{table}

\section{Discussion and conclusion}
\label{sec:discu&conclu}
This work has presented IterMask3D, an unsupervised anomaly segmentation framework developed for 3D brain MRI. By introducing an \textbf{iterative spatial mask refinement} strategy combined with \textbf{high-frequency structural guidance}, IterMask3D effectively detects reconstruction errors in anomalous regions while minimizing false positives in normal tissue. Additionally, we proposed a \textbf{subject-specific thresholding strategy} 
enabling adaptive anomaly detection, alleviating dependence on thresholds entirely determined using in-distribution validation datasets. 
Through extensive evaluations on multiple datasets, including synthetic and real-world imaging artifacts, as well as multiple brain pathologies across MRI sequences, IterMask3D consistently demonstrated superior or competitive performance compared to previous methods.

The promising performance across extensive evaluations underscores the potential of our method. However, several challenges remain—many of which are common in the current literature—and these point out valuable directions for future work.
\
A common challenge for reconstruction-based anomaly detection methods is the reduced sensitivity to anomalies that produce only subtle reconstruction errors. Such examples are small artifacts or patterns with average (gray) intensities.
An inherent challenge that exacerbates this issue is potential domain gap between training and testing datasets (differences in their imaging characteristics). It is often impractical to train the model on data obtained from the exact same imaging center or scanner as the testing data. Although our iterative mask refinement approach progressively reduces this domain gap by iteratively exposing to the model larger regions from the test image at inference time, anomalies that generate reconstruction errors similar to those caused by domain shifts can be mistakenly considered normal. Consequently, these anomalies may become unmasked and remain undetected during the iterative process.


Another challenge is that medical imaging data, especially MRI, differ significantly from natural images in terms of intensity range variability. While natural images typically have a fixed intensity range (e.g. 0–255), intensity ranges in MRI images exhibit substantial variability across datasets, scanners and acquisition protocols, as MRI intensity units are not standardized. Large intensity shift due to this between the training and test distribution can lead to suboptimal behavior of anomaly segmentation methods. A related challenge is the normalization of medical image intensities, which aims to assign comparable intensity ranges to the same tissue  across images, so that intensity shifts do not negatively affect reconstruction accuracy and anomaly detection. MRI intensity normalization is still an active field of research and current normalization methods can be severely affected by the presence of anomalies in images. Consequently, suboptimal intensity normalization can lead to further challenges for anomaly detection or segmentation.


Despite these challenges, our approach, IterMask3D, makes a meaningful contribution to the field by considering the sensitivity–precision trade-off and introducing iterative mask shrinking as a means to mitigate it in 3D anomaly detection. 
Results demonstrate particularly promising performance for detection of imaging artifacts which could be of practical usefulness for quality-assurance in imaging workflows. 
Results on unsupervised pathology segmentation, although achieving state-of-the-art for unsupervised segmentation, they still have room for improvement towards levels achievable by supervised segmentation methods. The improvement demonstrated by this method, however, shows there is substantial potential in unsupervised anomaly segmentation that is still untapped. Given that this category of methods could unlock applications for which supervised methods is less appropriate, such as segmentation of any type of brain lesion or incidental findings, we hope this study will motivate further future research in this direction. 

\section*{Acknowledgments}

For this work, ZL is supported by a scholarship provided by the EPSRC Doctoral Training Partnerships programme [EP/W524311/1]. YI is supported by the EPSRC Centre for Doctoral Training in Health Data Science (EP/S02428X/1). The authors also acknowledge UKRI grant reference [EP/X0

\noindent40186/1] and EPSRC grant [EP/T028572/1]. NLV gratefully acknowledges support from the NIHR Oxford Health Biomedical Research Centre [NIHR203316]. The views expressed are those of the author(s) and not necessarily those of the NIHR or the Department of Health and Social Care. The Wellcome Centre for Integrative Neuroimaging is supported by core funding from the Wellcome Trust (203139/Z/16/Z and 203139/A/16/Z). The authors also acknowledge the use of the University of Oxford Advanced Research Computing (ARC) facility in carrying out this work(http://dx.doi.org/10.5281/zenodo.22558).

Data used in preparation of this article were obtained from the Alzheimer’s Disease Neuroimaging Initiative (ADNI) database (adni.loni.usc.edu). As such, the investigators within the ADNI contributed to the design and implementation of ADNI and/or provided data but did not participate in analysis or writing of this report. A complete listing of ADNI investigators can be found at: \url{https://adni.loni.usc.edu/wp-content/uploads/how_to_apply/ADNI_Acknowledgement_List.pdf}. Data collection and sharing for this project were funded by the Alzheimer's Disease Neuroimaging Initiative (ADNI) (National Institutes of Health Grant U01 AG024904) and DOD ADNI (Department of Defense award number W81XWH-12-2-0012). ADNI is funded by the National Institute on Aging, the National Institute of Biomedical Imaging and Bioengineering, and through generous contributions from the following: AbbVie, Alzheimer’s Association, Alzheimer’s Drug Discovery Foundation, Araclon Biotech, BioClinica, Inc., Biogen, Bristol-Myers Squibb Company, CereSpir, Inc., Cogstate, Eisai Inc., Elan Pharmaceuticals, Inc., Eli Lilly and Company, EuroImmun, F. Hoffmann-La Roche Ltd and its affiliated company Genentech, Inc., Fujirebio, GE Healthcare, IXICO Ltd., Janssen Alzheimer Immunotherapy Research \& Development, LLC., Johnson \& Johnson Pharmaceutical Research \& Development LLC., Lumosity, Lundbeck, Merck \& Co., Inc., Meso Scale Diagnostics, LLC., NeuroRx Research, Neurotrack Technologies, Novartis Pharmaceuticals Corporation, Pfizer Inc., Piramal Imaging, Servier, Takeda Pharmaceutical Company, and Transition Therapeutics. The Canadian Institutes of Health Research provide funds to support ADNI clinical sites in Canada. Private sector contributions are facilitated by the Foundation for the National Institutes of Health (\url{www.fnih.org}). The grantee organization is the Northern California Institute for Research and Education, and the study is coordinated by the Alzheimer’s Therapeutic Research Institute at the University of Southern California. ADNI data are disseminated by the Laboratory for Neuroimaging at the University of Southern California.

Data were provided in part by OASIS [OASIS-3: Longitudinal Multimodal Neuroimaging: Principal Investigators: T. Benzinger, D. Marcus, J. Morris; NIH P30 AG066444, P50 AG00561, P30 NS09857781, P01 AG026276, P01 AG003991, R01 AG043434, UL1 TR000448, R01 EB009352. AV-45 doses were provided by Avid Radiopharmaceuticals, a wholly owned subsidiary of Eli Lilly.]

For the purpose of open access, the authors have applied a creative commons attribution (CC BY) licence (where permitted by UKRI, ‘open government licence’ or ‘creative commons attribution no-derivatives (CC BY-ND) licence’ may be stated instead) to any author accepted manuscript version arising.

During the preparation of this work the authors used ChatGPT to improve writing and grammar. After using this tool, the authors reviewed and edited the content as needed and take full responsibility for the content of the publication.
\clearpage

\bibliographystyle{elsarticle-harv} 
\bibliography{example}






\end{document}